\documentclass[10pt,journal,compsoc]{IEEEtran}
\ifCLASSOPTIONcompsoc
  \usepackage[nocompress]{cite}
\else
  \usepackage{cite}
\fi
\ifCLASSINFOpdf
\else
\fi
\usepackage{array}
\usepackage{orcidlink}

\usepackage{times}
\usepackage{epsfig}
\usepackage{graphicx}
\usepackage{amsmath}
\usepackage{amssymb}
\usepackage{enumitem}
\usepackage{caption}
\usepackage{xcolor}
\usepackage{soul}
\usepackage{enumitem}
\usepackage{multirow,multicol}
\usepackage{amsmath,graphicx}
\usepackage{subcaption}
\usepackage[flushleft]{threeparttable}
\usepackage{arydshln}

\usepackage{etoolbox}
\usepackage{booktabs}

\usepackage{scalerel}
\usepackage{tikz}
\usetikzlibrary{svg.path}

\definecolor{orcidlogocol}{HTML}{A6CE39}
\tikzset{
    orcidlogo/.pic={
        \fill[orcidlogocol] svg{M256,128c0,70.7-57.3,128-128,128C57.3,256,0,198.7,0,128C0,57.3,57.3,0,128,0C198.7,0,256,57.3,256,128z};
        \fill[white] svg{M86.3,186.2H70.9V79.1h15.4v48.4V186.2z}
        svg{M108.9,79.1h41.6c39.6,0,57,28.3,57,53.6c0,27.5-21.5,53.6-56.8,53.6h-41.8V79.1z M124.3,172.4h24.5c34.9,0,42.9-26.5,42.9-39.7c0-21.5-13.7-39.7-43.7-39.7h-23.7V172.4z}
        svg{M88.7,56.8c0,5.5-4.5,10.1-10.1,10.1c-5.6,0-10.1-4.6-10.1-10.1c0-5.6,4.5-10.1,10.1-10.1C84.2,46.7,88.7,51.3,88.7,56.8z};
    }
}

\newcommand\orcidicon[1]{\href{https://orcid.org/#1}{\mbox{\scalerel*{
                \begin{tikzpicture}[yscale=-1,transform shape]
                \pic{orcidlogo};
                \end{tikzpicture}
            }{|}}}}

\usepackage{hyperref} 

\begin{document}
%
\title{CATFace: Cross-Attribute-Guided Transformer with Self-Attention Distillation for Low-Quality Face Recognition}

\author{Niloufar~Alipour~Talemi$^{\textsuperscript{\orcidicon{0009-0000-6881-3671}}}$\,, Hossein~Kashiani$^{\textsuperscript{\orcidicon{0000-0001-8338-9987}}}$\,, and Nasser M. Nasrabadi$^{\textsuperscript{\orcidicon{0000-0001-8730-627X}}}$,~\IEEEmembership{Fellow,~IEEE}}

\markboth{TO APPEAR IN IEEE TRANSACTIONS ON BIOMETRICS, IDENTITY AND BEHAVIOR (T-BIOM)}%
{Shell \MakeLowercase{\textit{et al.}}: Bare Demo of IEEEtran.cls for Biometrics Council Journals}
%



\IEEEtitleabstractindextext{%
\begin{abstract}
Although face recognition (FR) has achieved great success in recent years, it is still challenging to accurately recognize faces in low-quality images due to the obscured facial details. Nevertheless, it is often feasible to make predictions about specific soft biometric (SB) attributes, such as gender, and baldness even in dealing with low-quality images. In this paper, we propose a novel multi-branch neural network that leverages SB attribute information to boost the performance of FR. To this end, we propose a cross-attribute-guided transformer fusion (CATF) module that effectively captures the long-range dependencies and relationships between FR and SB feature representations. The synergy created by the reciprocal flow of information in the dual cross-attention operations of the proposed CATF module enhances the performance of FR. Furthermore, we introduce a novel self-attention distillation framework that effectively highlights crucial facial regions, such as landmarks by aligning low-quality images with those of their high-quality counterparts in the feature space. The proposed self-attention distillation regularizes our network to learn a unified quality-invariant feature representation in unconstrained environments. We conduct extensive experiments on various FR benchmarks varying in quality. Experimental results demonstrate the superiority of our FR method compared to state-of-the-art FR studies.
\end{abstract}

}

\maketitle

\IEEEdisplaynontitleabstractindextext

%
\IEEEpeerreviewmaketitle

\IEEEraisesectionheading{\section{Introduction}\label{sec:introduction}}

%
%
%
%
\IEEEPARstart{F}{ace} recognition (FR) has been one of the most popular fields in computer vision due to its wide range of applications in military, public security, and daily life~\cite{khallaf2021classification}. In the realm of FR, there has been notable progress in recent years with the emergence of advanced network architectures \cite{szegedy2015going, he2016deep}. Alongside these advancements, the field has witnessed the introduction of various designs of loss functions \cite{wang2017normface,deng2019arcface,liu2017sphereface,wang2018cosface,meng2021magface, kim2022adaface, wang2018cosface} which have played a significant role in enhancing FR performance. Despite all these advancements, it has still been challenging to preserve the high performance of FR methods in unconstrained environments. The majority of FR training datasets \cite{cao2018vggface2,yi2014learning,guo2016ms} consist of high-quality images that differ significantly from real-world environments. This becomes evident when considering images captured by surveillance cameras \cite{maze2018iarpa}, which present challenging attributes like sensor noise, low resolution, motion blur, and turbulence effect, among others. As a result, when FR models trained on constrained datasets are applied to real-world scenarios, the models' accuracy suffers a significant drop. On the other hand, collecting a large-scale unconstrained face dataset with large variations needs manual labeling, which is time-consuming and costly to provide.

\begin{figure}[!ttbp]
    \centering 
    \includegraphics[scale=.15]{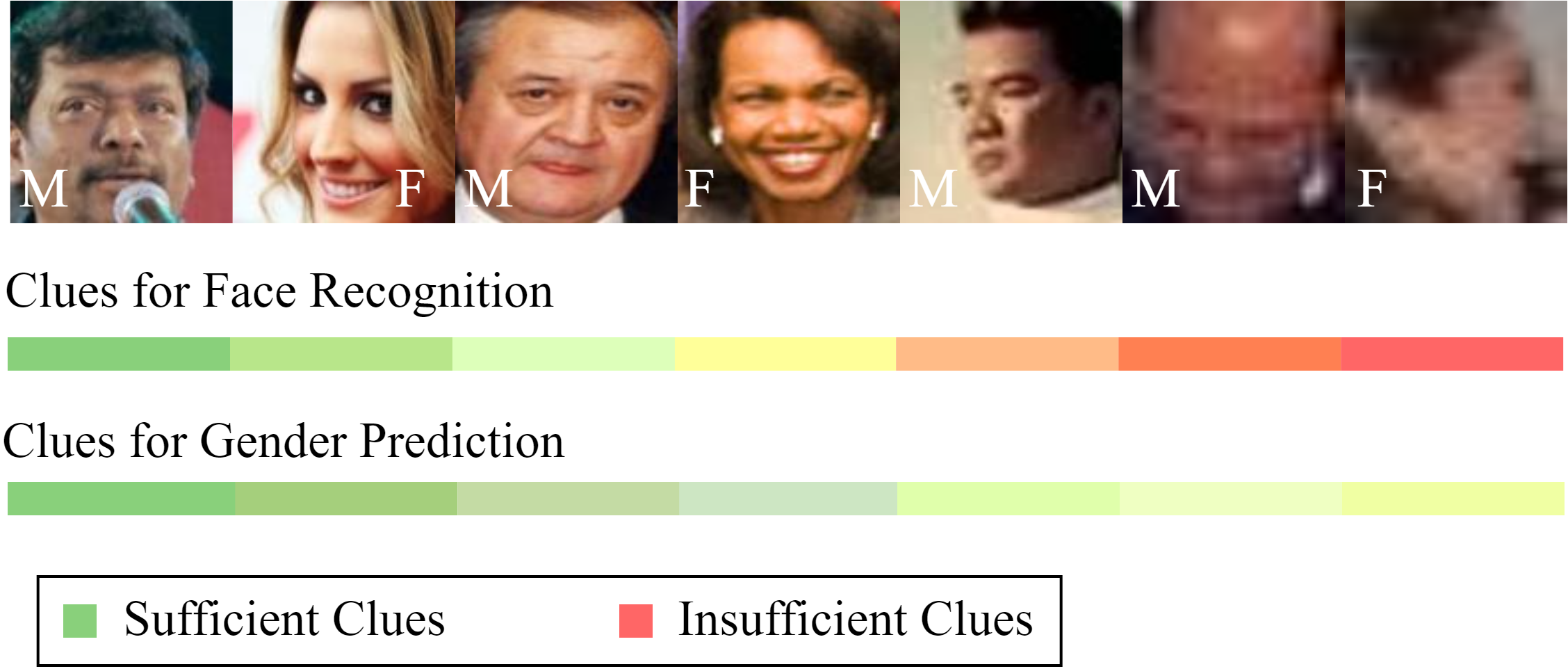}
    \caption{Examples of face images with different degrees of degradation in various real-world FR
benchmarks. In some images, the identity of the person is not easily recognizable due to the lack of some clues that are essential for FR. However, the gender of a person can still be inferred from those images. Therefore, leveraging some SB attributes like gender can enhance FR performance in challenging conditions. Note that M and F stand for male and female, respectively.}
    \label{fig:overview}

\end{figure}

Recently, several alternative approaches have been proposed to fill the gap between the semi-constrained training datasets and unconstrained testing sets. Some of these studies involve super resolution-based techniques \cite{tran2017disentangled, singh2022towards, chen2020identity}, which aim to reconstruct high-resolution images from low-resolution ones and then feed them to a FR model. Moreover, with the advent of Generative Adversarial Networks (GANs) \cite{goodfellow2014generative, NIPS2016_502e4a16, Karras_2019_CVPR} several GAN-based frontalization approaches \cite{tu2021joint,towards-large-pose-face-frontalization-in-the-wild} have been proposed to handle faces with extreme poses. However, these approaches primarily address specific image variations, resulting in limited generalizability across diverse conditions. Another concern is that during the inference step, performing such preprocessing methods may lead to significantly high computational cost compared to the recognition network itself. Furthermore, despite the significant advancements in GAN models, preserving identity in particular for cases with extreme poses has been still a challenging problem \cite{yin2021superfront, cao2020towards}.

Looking at the FR task from an alternative perspective, humans inherently analyze facial attributes to discern the identity of a person. This observation can bring up the hypothesis that utilizing facial attributes can improve the performance of FR in challenging cases. Fig. \ref{fig:overview} shows that even in the case of low-quality images, it is often feasible to predict certain soft biometric (SB) attributes like gender, while accurately identifying the exact identity proves to be a challenging task. Therefore, the incorporation of SB information into FR can help the network to recognize the identities more accurately. Inspired by this observation, in this work, we employ facial attributes including gender, and baldness, which remain consistent across different scenarios such as varying illuminations and poses to raise the performance of FR. To this end, we propose a multi-head neural network that not only predicts SB attributes and recognizes identities simultaneously but also employs a novel cross-attribute-guided transformer fusion (CATF) module to effectively integrate feature representation of the SB information into FR features. This module initially enables the synergistic fusion of the SB and FR feature representations using dual cross-attention operations. Subsequently, it conducts feature fusion across global spatial tokens, where each channel is treated as an independent token. By regarding each channel token as an abstract representation of the entire image, the fusion process naturally captures global interactions and representations. Therefore, our proposed fusion module concentrates on the most pertinent areas within both the SB and FR feature representations and enhances the integration of distinctive facial details with SB cues. 

To further leverage the advantages of using SB information as an auxiliary modality, we train our proposed multi-branch network using a novel knowledge distillation (KD) based approach. This approach enables our SB prediction branch to exhibit robustness in challenging cases, thereby improving the overall FR accuracy. The idea of utilizing KD, which is a teacher-student-based framework, originated from the observation that although crucial visual details are missing in low-quality images, in many cases humans can still roughly determine an object’s regions in such images based on prior knowledge learned from previously viewed corresponding high-quality images \cite{jin2001face}. Thus, as in low-quality faces, features from detailed parts of a face may not be captured, our KD approach tries to transfer prior knowledge from the high-quality images to the low-quality ones. Furthermore, to eliminate the necessity of a pre-trained teacher network, we adopt a self-KD method that involves training a single network in a progressive manner to distill its own knowledge \cite{ji2021refine}.

In most self-KD-based approaches, the primary focus lies in minimizing the distance between the feature maps or soft targets of the networks. However, in this work, we adopt a different perspective by leveraging attention values derived from the network's feature representation. When attention is applied to the feature representations, the importance of essential regions, such as face landmarks, is heightened. This leads to the distillation of more significant information that effectively contributes to the FR task. As self-attention modules are the fundamentally important parts of our proposed CATF module which is based on transformers, we emulate the self-attention mechanism in our KD approach. To be more specific, we distill the knowledge from the high-quality self-attention components to their corresponding low-quality counterparts. Furthermore, considering the positive correlation between feature norm and image quality in recent studies \cite{meng2021magface, kim2022adaface}, our approach centers on aligning only the directional component of attention maps rather than their magnitude. Consequently, we formulate our distillation loss using cosine similarity, which enables us to capture the angular relationship between feature vectors and enhances the discriminative power of the model.

Most previous KD-based FR studies \cite{ge2020efficient, ge2018low} manually create low-resolution images to train the student network while low-resolution is merely one probable characteristic that unconstrained images may have. Hence, in this work, to fully exploit the advantage of self-KD, we augment the training dataset with diverse properties encountered in real-world scenarios. These properties include atmospheric turbulence, improper illuminations, motion blur, and various styles. By incorporating such variations, we aim to enhance the robustness and generalization ability of the network, enabling it to effectively recognize faces under challenging conditions commonly encountered in everyday life. This comprehensive augmentation approach moves beyond the limited scope of low-resolution images and provides a more realistic and representative training environment for KD-based FR systems. Some examples of the augmented data that we utilize in our KD-based approach are illustrated in Fig. \ref{fig:aug}. 

In summary, the key contributions of this paper are summarized as follows:
\begin{enumerate}
\item We propose a novel multi-branch neural network to tackle the challenge of FR in low-quality images. By leveraging certain facial attributes from the SB branch, we enhance the performance of the dedicated FR branch.
\item We present a cross-attribute-guided transformer fusion (CATF) module which effectively captures and incorporates long-range dependencies, enabling a comprehensive understanding of the intricate relationships between FR and SB feature representations.
\item To raise the robustness of our proposed network against low-quality images, we propose a novel KD-based training approach. We prioritize important areas like facial landmarks by distilling self-attention values, outperforming other KD-based methods.
\item  Extensive experiments with diverse datasets, including manual and real-world low-quality images, strongly support the enhanced performance of FR through employing SB information with the novel KD-based training approach and the proposed CATF module. 

\begin{figure}[t]
    \centering 
    \includegraphics[scale=0.087]{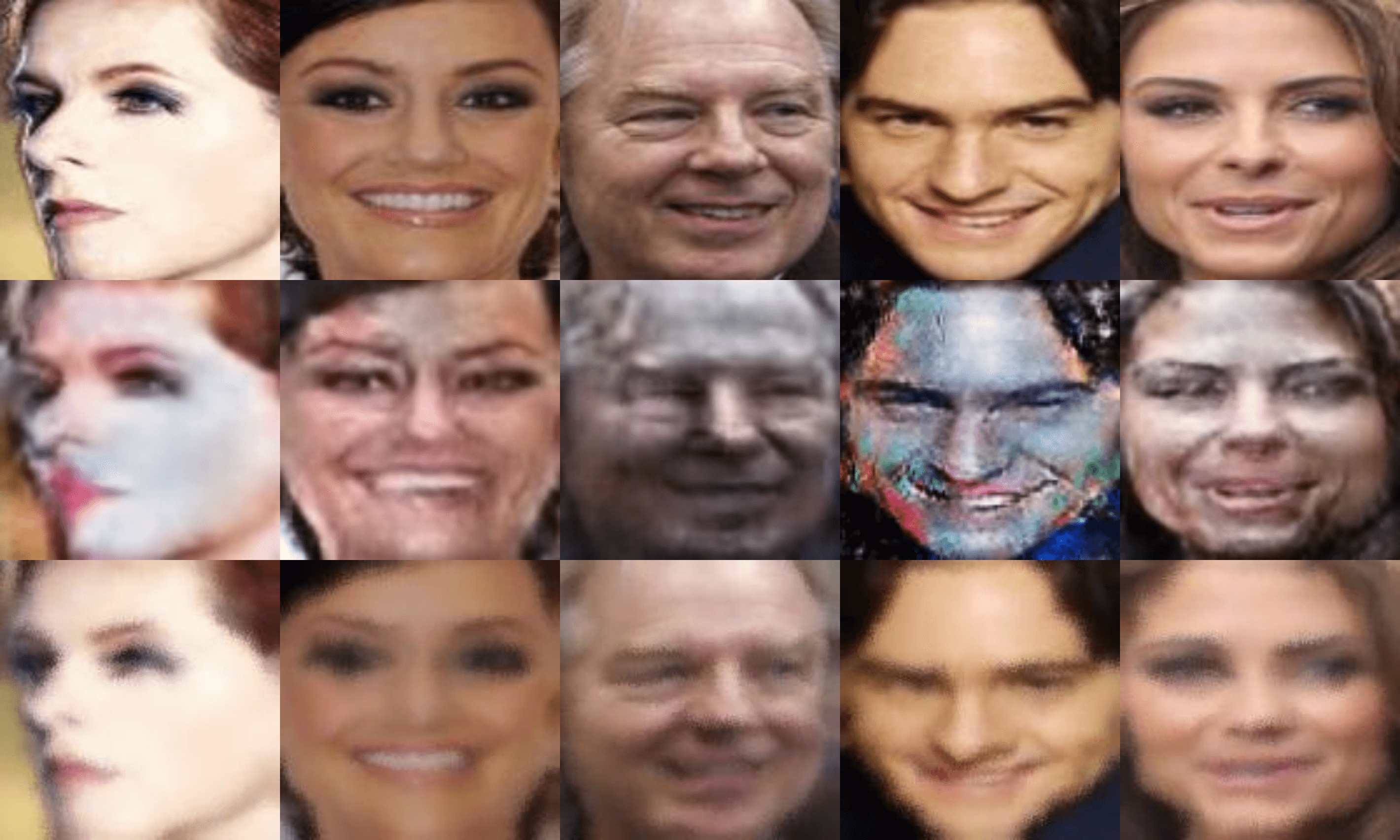}
    \caption{Samples of augmented data. The first row shows original images from the CelebA dataset. The second and the third rows depict low-quality versions of the original data which are generated by controllable face synthesis GAN and atmospheric turbulence simulator, respectively.}
    \label{fig:aug}
    
\end{figure}

\end{enumerate}

\section{Related Works}
\subsection{Data Augmentation} \label{augmen}
One of the most significant challenges of FR methods is low-quality images. Low-quality images can result from several factors such as inherent camera noise, atmospheric turbulence, and improper illuminations which lead to significant performance degradation. 
One promising approach to address this challenge is data augmentation, which involves creating additional training data by artificially modifying the existing dataset. By exposing the network to a wide range of image variations during training, it can learn to better cope with different types of distortions that may be present in real-world scenarios. Therefore, in this work, we augment our training data with different styles and attributes to enhance the robustness and generalization ability of our proposed model.

\begin{itemize} 
\item \textbf{Controllable Face Synthesis.}
In 2014, Goodfellow et al. introduced GANs \cite{goodfellow2014generative} to synthesize realistic data samples based on a pair of neural networks, namely a generator and a discriminator. The core concept of GANs involves training the two networks in an adversarial manner, where the generator learns to produce fake samples that are indistinguishable from real ones, while the discriminator is trained to differentiate between real and fake samples. In recent years, many GAN-based face image generation models have been proposed to synthesize face images with desired properties \cite{roich2022pivotal, xiao2018elegant}. However, many of these models primarily focus on face editing tasks, such as face aging or transferring diverse expressions and poses from a given face image to a target one. While these approaches are valuable for such tasks, they are not particularly beneficial for enhancing the performance of FR in low-quality scenarios. In this regard, to create realistic useful face images, we use the model proposed in \cite{liu2022controllable} which is a novel face synthesis model that can generate face images similar to the distribution of a target dataset through learning a linear subspace in the style latent space. Therefore, with an unlabeled target dataset including our desired characteristics such as motion blur, inherent sensor noise, and low resolution, we can generate face images containing all such attributes. Employing such realistic low-quality images through our proposed KD approach makes our multi-branch network robust against the different characteristics of the unconstrained scenarios.

\item \textbf{Atmospheric Turbulence Simulation.} 
The effects of atmospheric turbulence on long-distance imaging applications, particularly in areas such as surveillance, are substantial. The fluctuation in the refractive index of air caused by atmospheric turbulence leads to variations in the path of light through the atmosphere, resulting in distortions in the captured images \cite{mao2021accelerating, robbins2022effect}. These distortions significantly degrade the image quality, thereby posing challenges in extracting useful information from the affected images. Therefore, we consider the atmospheric turbulence effect in the training process. To generate atmospheric turbulence, we use a Phase-to-Space simulator which is proposed in \cite{mao2021accelerating}. This simulator is based on a novel concept called the phase-to-space (P2S) transform, which converts the phase representation of the turbulence to the spatial representation of the image. The P2S transform is implemented by a light-weight neural network that learns the basis functions for the spatially varying convolution from the known turbulence statistics models. By using the P2S transform, the simulation can be significantly accelerated compared to the conventional split-step method, while preserving the essential turbulence statistics.

\end{itemize}
\subsection{Multi-task Learning for Face Analysis}

Multi-task learning (MTL) is a strategy that simultaneously optimizes several relevant tasks to enhance the generalization performance through an inductive transfer mechanism. The concept of MTL can be traced back to the 1990s \cite{caruana1997multitask} which involves leveraging a single neural network to perform multiple related tasks. Following the advent of deep neural networks \cite{krizhevsky2012imagenet}, MTL has been applied in many computer vision tasks such as medical image analysis \cite{app11094247,chen2019multi}, object detection \cite{khattar2021cross, ZHANG202165}, and facial attribute recognition \cite{ranjan2017hyperface, ranjan2017all}. Here, we concentrate on MLT approaches for face analysis. Recently, several methods have incorporated the MTL framework for face-related tasks. Levi et al. \cite{levi2015age} used the MTL framework to simultaneously perform age and gender prediction from face images. Also, HyperFace \cite{ranjan2017hyperface} is a multi-task learning algorithm that operates on the fused intermediate features to predict facial attributes and to do some other face-related tasks. In All-In-One \cite{ranjan2017all}, authors took advantage of MTL for FR in addition to facial attribute prediction which led to considerable improvement in FR for challenging unconstrained environments. However, existing MTL frameworks, do not directly leverage attribute information to enhance the performance of a FR task whereas intrinsically humans analyze facial attributes to recognize identities. In this regard, we propose a new multi-branch neural network that simultaneously performs SB prediction as an auxiliary modality and FR as the main task. To boost the discriminative ability of the FR branch, we integrate SB information with FR feature representation through an attentional module that is capable of learning complex relationships between input features. Moreover, we rely on facial attributes that remain consistent across different images of the same identity. For instance, attributes such as gender and eye shape exhibit consistency across varying illuminations or poses, while others like hair color may vary within different images of the same individual. 
\subsection{Knowledge Distillation}
The concept of KD was first introduced by Hinton et al. in 2015 \cite{hinton2015distilling}. The basic idea is transferring knowledge from a large neural network (teacher) to a smaller neural network (student) by minimizing the Kullback-Leibler (KL) divergence of soft class probabilities between them. After that, several variants of distillation methods have been suggested to leverage the insights provided by the teacher network more effectively. For instance, many feature-based KD methods have been proposed that focus on distilling intermediate representations from the teacher model into the student network \cite{yim2017gift, park2019relational}. An alternative approach known as self-distillation involves training a student network using its own knowledge, without the need for a separate teacher network. This approach aims to improve the efficiency, generality, and transferability of the learned knowledge. For instance,  Zhang et al. in \cite{zhang2019your} proposed to first divide the network into several sections and then squeeze the knowledge in the deeper portion of the network into the shallow ones. There are also many augmentation-based strategies for self-distillation approaches \cite{xu2019data, yun2020regularizing}. In the field of FR, researchers have also investigated the application of KD methods to improve network performance through the use of augmentation techniques, especially for challenging scenarios \cite{shin2023enhancing}. Shin et al. \cite{shin2023enhancing} utilized manually created low-resolution images to train the student network. However, our work takes a different approach by utilizing various synthesis methods to generate realistic face images that exhibit the characteristic challenges observed in real-world scenarios, instead of relying on simple down-sampling. Furthermore, we go beyond typical methods that use network outputs or feature maps of different layers to transfer knowledge from high-quality images to low-quality ones. Instead, we leverage attention distillation to gain enhanced guidance from the self-attention mechanism by effectively identifying the crucial knowledge embedded within high-quality images. In addition to what is distilled, the distance metric for measuring distillation is also a critical factor influencing the model's performance (see {Section} \ref{cosine_norm}). Therefore, in contrast to most KD-based methods \cite{liu2022cross, yun2020regularizing} that employ general distance metrics like KL divergence or L2-distance, we tailor our distillation loss to our specific application which is FR.

\subsection{Vision Transformer}

Vision transformer (ViT) \cite{dosovitskiyimage} is a novel neural network architecture that adapts the transformer model, originally developed for natural language processing, to vision processing tasks such as image recognition. ViT relies on the self-attention mechanism as a key component.
This enables ViT to effectively attend to different parts of the input image and captures the interactions and long-range dependencies among pixels or patches within visual data. By leveraging self-attention, ViT can recognize complex patterns and features in the input images, thereby achieving state-of-the-art (SoTA) performance in large-scale image classification tasks. Inspired by the success of the seminal ViT, researchers have recognized the potential of the ViT architecture in various computer vision tasks like FR \cite{luo2022memory,zhong2021face,george2022prepended}. The potential of the vanilla ViT architecture for FR is explored in \cite{zhong2021face}. This study finds that although the transformer network encounters challenges when working with smaller databases, it exhibits promising performance when trained on larger datasets. Su et al. \cite{su2023hybrid} introduce the atomic tokens and holistic tokens in the transformer encoder to capture the attentive relationship between facial regions and learn discriminative hybrid tokens to boost FR performance. Unlike the studies in FR that rely on the vanilla transformer architecture as a feature extractor, we exploit the advances in recent ViT studies to tailor the CATF module that can capture both global and local dependencies in face images between the FR and SB tasks. By incorporating the CATF module into our model, we enable a holistic understanding of the relationships between the FR and SB tasks, facilitating effective feature fusion. This would make our model achieve SoTA performance in the FR task which is our main goal.

\begin{figure*}[t]
    \centering 
    \includegraphics[scale=.145]{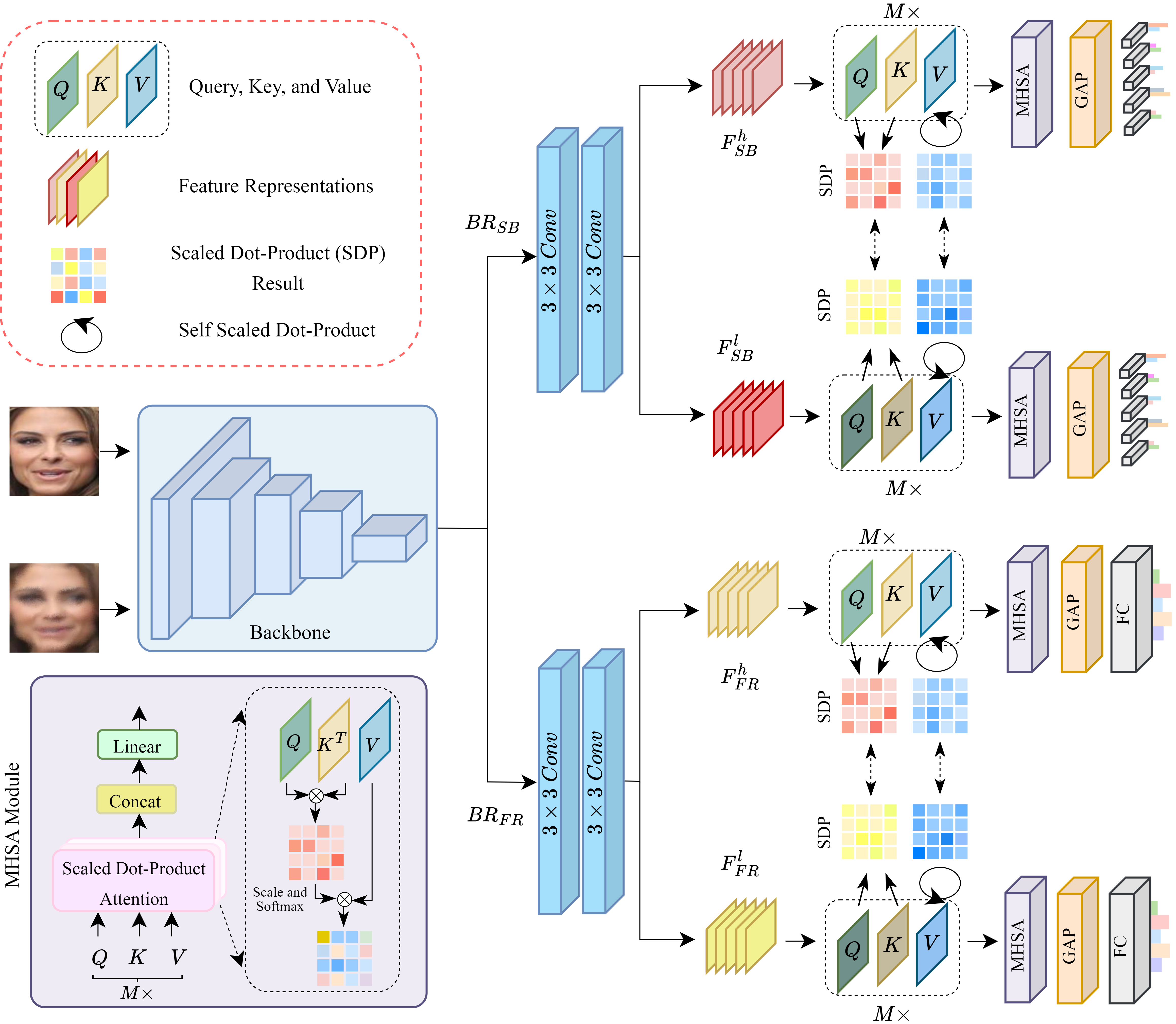}
   \caption{ Multi-branch neural network with self-attention distillation for FR and SB attribute prediction. Note that MHSA stands for multi-head self-attention module. This diagram shows the first step of our two-step training process. The $Br_{FR}$ and $Br_{SB}$ branches are jointly trained in the first step of the training process. In the second step, to enrich the FR feature representations, the SB and FR feature representations are fused together through the proposed CATF module (see Fig. \ref{fig:mbn}). It should be noted that the global average pooling (GAP) and the final fully connected (FC) layers are removed from each branch for the second step of the training process.} 
    \label{fig:Proposed Network}

\end{figure*}
\section{Proposed method} \label{Proposed method}

The proposed architecture, illustrated in Fig. \ref{fig:Proposed Network}, comprises two branches, namely the $Br_{FR}$ and $Br_{SB}$, both sharing a ResNet-based backbone and dedicated convolutional layers. The $Br_{FR}$ is dedicated to FR, while the $Br_{SB}$ is intended for SB prediction. To enhance the performance of the network against low-quality images, we employ a self-distillation approach during the simultaneous training of both branches. By leveraging this approach, our network is capable of extracting and distilling valuable information from high-quality samples, thereby enhancing its ability to handle low-quality input images. This methodology showcases a promising direction for addressing the challenges posed by low-quality image inputs in FR and SB prediction tasks. Additionally, as shown in Fig. \ref{fig:mbn}, upon training the multi-branch network,  we employ a novel attention mechanism to integrate the SB information into the FR feature representation. This integration effectively enriches the FR embedding, ultimately improving the model's ability to identify challenging face images.

\subsection{Self-Distilled Multi-Branch Network} \label{Self-Distilled Multi-Bra}
\subsubsection{Multi-Branch Network}

For FR, as it is shown in Fig. \ref{fig:Proposed Network}, there exist dedicated convolutional layers in addition to the backbone. The FR branch concludes with a softmax layer, the dimensions of which are determined by the number of classes within the training dataset. Thus, this branch is intrinsically considered for face identification which is basically a classification task based on identities. Conventional softmax loss of a sample $x_i$ can be expressed as:
\begin{equation} \label{eq_softmax}
L(x_i) = -\log \dfrac {exp({W_{y_{i}}.z_{i}+b_{y_{i}}})}{\Sigma_{j}^{N_{c}} exp({W_{j}.z_{j}+b_{j})}},
\end{equation} 

\noindent where $W_{j}$ represents the $j$-th column of the last fully connected layer’s weight, $W\in\mathbb{R}^{d\times N_{c}}$. The face embedding of sample $x_i$ and its ground truth identity are shown by $z_{i}\in\mathbb{R}^d$ and $y_{i}$, respectively. $N_{c}$\: and $b_{j}$ indicate the number of classes and the bias term for the $j$-th class, respectively. In the context of training FR models, the features obtained from a simple softmax loss often fail to exhibit sufficient discriminative power. To address this limitation, the prevalent approach is to employ a margin-based softmax loss function. This loss function leads to the minimization of intra-class compactness, ensuring samples within the same identity cluster closely together, while simultaneously maximizing inter-class dispersion, enabling better separability between samples from different identity classes. In margin-based loss functions, for simplicity, the bias term is fixed to 0, and also the inner product of features and weights is considered as $\|W_{j}\| \|z_{i}\|$ $cos \theta_{j}$ \cite{liu2016large}. $\theta_{j}$ corresponds to the angle between $z_{i}$ and $W_{j}$. Assuming $||W_{j}||$ to be equal to 1 and $z_{i}$ is rescaled with $s$ during training, margin-based loss functions can be expressed as follows:

\begin{equation} \label{margin_based_softmax}
{L}_{\, AdaFace}(x_i) = -\log \dfrac {exp(f(\theta_{y_{i}}, m))}{exp(f(\theta_{y_{i}}, m)) +\Sigma_{j\neq y_{i}} exp(scos\theta_{j})},
\end{equation}

\noindent where $m$ is a scalar hyper-parameter referred to as the margin, and $f$ is a margin function. To achieve better convergence, many margin-based loss functions have been introduced where $f(\theta_{y_{i}}, m)$ is the only distinguishing factor among them. AdaFace\cite{kim2022adaface} is one of the recent SoTA margin-based loss functions that emphasizes samples of different difficulties based on their image quality. It approximates the image quality with feature norms. For high norms, it emphasizes samples away from the boundary, and for low norms, it emphasizes samples near the boundary. In this work, to train the FR branch, we utilize AdaFace loss function in which the margin function is defined as follows;

\begin{equation} \label{adaface_loss}
f(\theta_{y_{i}}, m) = 
\begin{cases}
      s(cos(\theta_{j}+g_{angle})-g_{add}) & \text{if $j = y_{i}$}\\
      s \, cos\theta_{j} & \text{if $j \neq y_{i}$}\\
    \end{cases}       
,
\end{equation}

\begin{equation} \label{adaface_con}
 g_{angle}= -m \cdot \widehat{\|z_{i}\|}, \;  g_{add}= m \cdot \widehat{\|z_{i}\|} + m,
\end{equation}

\begin{equation} \label{adaface_con2}
\widehat{\|z_{i}\|} = \Biggl \lfloor(\dfrac {\|z_{i}\| - \mu_{z}} {\dfrac{\sigma_{z}}{h}})\Biggr \rceil_{-1}^{1} ,
\end{equation}  

\noindent where $\|z_{i}\|$ measures the quality of a sample $i$, and $\widehat{\|z_{i}\|}$ is the normalized quality using mean ($\mu_{z}$) and standard deviation ($\sigma_{z}$) of all $z_{i}$ within a batch. It should be noted that over the test time when we are presented with arbitrary pairs of images for comparison (e.g. $I_{1}$ and $I_{2}$), the cosine similarity metric between them ($\dfrac{z_{1} \cdot z_{2}} {\|z_{1}\| \|z_{2}\|}$) determines whether they belong to the same identity or not. 

The architecture of the $Br_{SB}$ is similar to the $Br_{FR}$. During the training process, we employ binary classifiers dedicated to predicting different facial attributes, with each classifier equipped with its respective cross-entropy loss function. The total classification loss for the $Br_{SB}$ is given by:
\begin{equation} \label{eq1}
L_{SB} = \sum_{i=1}^{n}{\lambda_{a_{i}}L_{a_{i}}},
\end{equation}

\noindent where each $L_{a_{i}}$ represents a loss for each individual attribute and $\lambda_{a_{i}}$
is the loss-weight corresponding to the attribute $a_{i}$. Also, $n$ denotes the number of SB attributes in $Br_{SB}$. For each attribute, $L_{a_{i}}$ is computed as:
\begin{equation} \label{eq2}
L_{a_{i}} = - \left( a_{i} \log(p_{a_{i}}) + \left(1-a_{i}\right) \log{\left(1-p_{a_{i}}\right)} \right),
\end{equation} 
\noindent where $p_{a_{i}}$ is the probability that the network computes for $a_{i}$.

\begin{figure*}[t]
    \centering 
    \includegraphics[scale=0.128]{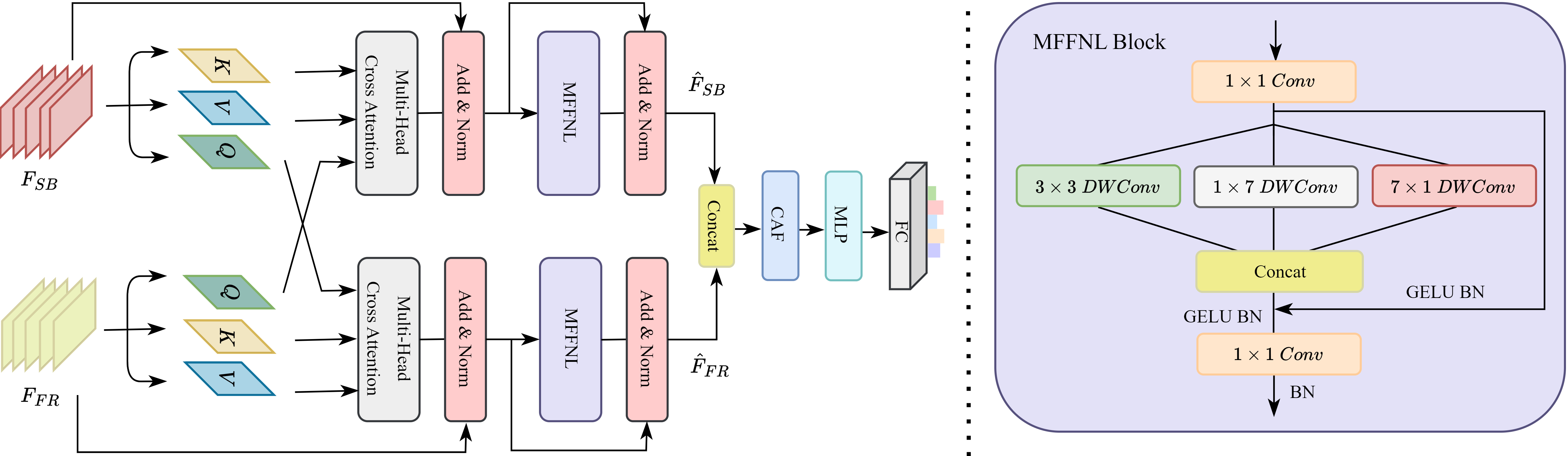}
    \caption{ Proposed cross-attribute-guided transformer fusion (CATF) module for FR. This diagram shows the second step of our two-step training process.}
    \label{fig:mbn}%
\end{figure*}
\subsubsection{Self-Attention Distillation}\label{Self_Attention_Distillation_proposed}
To enable our proposed multi-branch network to have a robust performance against low-quality images, we employ a self-distillation mechanism to distill information from high-quality images to their corresponding low-quality ones. In comparison with the most conventional KD-based methods that directly make feature maps close to each other, we focus on keeping the attention maps consistent between high-quality and low-quality samples in both the SB and FR branches. 
The inspiration behind our approach lies in the attention mechanism, which guides feature maps to focus on important regions. As key-points such as eyes and mouth are crucial for both FR and SB prediction, employing attention-based distillation can help the network to distill more informative features. 

To identify the most discriminative features, we apply self-attention to the last convolutional layer of both the SB and FR branches. Self-attention can capture global dependencies between different regions of the face image such as face landmarks. Furthermore, in our proposed method, SB and FR feature representations are integrated using a cross-attribute-guided transformer fusion (CATF) module in which self-attention is a principal block. As a result, to provide the best input for the fusion module which includes three key components (i.e., key, query, and value), we leverage the self-attention mechanism to teach the network key information from high-quality samples.

Let ${F}_{FR}^{h} \in \mathbb{R}^{H\times W\times C}$ and ${F}_{FR}^{l} \in \mathbb{R}^{H\times W\times C}$ be the feature maps of the last convolutional layer of the $BR_{FR}$ for a high-quality sample and its corresponding low-quality one, respectively, where $H$, $W$, and $C$ denote the height, width, and the channel number of each feature map. Similarly, we assume ${F}_{SB}^{h}\in \mathbb{R}^{H\times W\times C}$, and ${F}_{SB}^{l}\in \mathbb{R}^{H\times W\times C}$ for $BR_{SB}$. We first flatten them to ${F}_{FR}^{l}, {F}_{FR}^{h}, {F}_{SB}^{l}, {F}_{SB}^{h} \in \mathrm{R} ^{N\times C}$ ($N= H\times W$). Then, based on the self-attention mechanism, each feature map will be linearly projected to three learnable matrices: query matrices ($ \mathrm{Q}_{FR}^{l},\mathrm{Q}_{FR}^{h}, \mathrm{Q}_{SB}^{l}, \mathrm{Q}_{SB}^{h}  \in \mathrm{R} ^{ N \times C  }$), key matrices ($ \mathrm{K}_{FR}^{l},\mathrm{K}_{FR}^{h}, \mathrm{K}_{SB}^{l}, \mathrm{K}_{SB}^{h}  \in \mathrm{R} ^{ N \times C  }$), and value matrices ($ \mathrm{V}_{FR}^{l},\mathrm{V}_{FR}^{h}, \mathrm{V}_{SB}^{l}, \mathrm{V}_{SB}^{h}  \in \mathrm{R} ^{ N \times C  }$). Finally, the attention map will be computed as the dot product of each $\mathrm{Q}$ and its corresponding $\mathrm{K}$, as follows:

\begin{equation} \label{attention_map_1}
 \mathrm{A}_{FR}^{l} = \mathrm{Softmax}\Big( \frac{\mathrm{Q}_{FR}^{l}(\mathrm{K}_{FR}^{l})^\mathrm{T}}{\sqrt{C}}\Big) \,\mathrm{V}_{FR}^{l},
\end{equation}

\begin{equation} \label{attention_map_2}
 \mathrm{A}_{FR}^{h} = \mathrm{Softmax}\Big( \frac{\mathrm{Q}_{FR}^{h}(\mathrm{K}_{FR}^{h})^\mathrm{T}}{\sqrt{C}}\Big) \,\mathrm{V}_{FR}^{h},
\end{equation}

By following the same computational process, we can also compute $\mathrm{A}_{SB}^{l}$ and $\mathrm{A}_{SB}^{h}$.
We force the network to mimic not only attention maps of high-quality samples but also their corresponding value parameters. To minimize the differences between attention maps and also value parameters of high-quality and low-quality samples, we employ cosine similarity. Therefore, the distillation loss is computed as:
\begin{equation} \label{loss_distill_0}
L^{distill} = L_{FR}^{distill} + L_{SB}^{distill},
\end{equation}
\begin{equation} \label{loss_distill1}
L_{FR}^{distill} = 2-\langle\mathrm{A}_{FR}^{l}, \mathrm{A}_{FR}^{h}\rangle- \langle\mathrm{V}_{FR}^{l}, \mathrm{V}_{FR}^{h}\rangle,  
\end{equation}
\begin{equation} \label{loss_distill2}
L_{SB}^{distill} = 2-\langle\mathrm{A}_{SB}^{l}, \mathrm{A}_{SB}^{h}\rangle - \langle\mathrm{V}_{SB}^{l}, \mathrm{V}_{SB}^{h}\rangle,
\end{equation}

\noindent where $\langle . \rangle$ denotes the cosine similarity metric. The total loss for each branch is the weighted sum of the target task’s loss and the distillation loss.

\subsection{Cross-Attribute-Guided Transformer Fusion (CATF).}

To selectively focus on the most relevant regions in both the SB and FR feature representations and facilitate the fusion of discriminative facial information with SB cues, we employ a dual cross-attention operations in the CATF module (see Fig. \ref{fig:mbn}). The reciprocal flow of information in the dual cross-attention operations enables a synergistic fusion of the SB and FR feature representations, enhancing the overall performance of FR. The cross-attention operations also effectively capture long-range dependencies, providing a holistic understanding of the relationships between the FR and SB tasks for feature fusion. In addition, we propose a multi-scale feed-forward network with locality (MFFNL), and the channel-wise attentional fusion (CAF) block in the CATF module to further improve the fusion of discriminative facial information
with SB cues.
Given the feature representations of SB and FR as ${F}_{FR} \in \mathbb{R} ^{H\times W \times C}$ and ${F}_{SB} \in \mathbb{R} ^{H\times W \times C}$, we separately map ${F_{FR}}$, and ${F_{SB}}$ to three learnable matrices: query matrices ($ \mathrm{Q}_{FR},\mathrm{Q}_{SB}  \in \mathbb{R} ^{ N \times C  }$), key matrices ($ \mathrm{K}_{FR},\mathrm{K}_{SB} \in \mathbb{R} ^{ N \times C } $), and value matrices ($ \mathrm{V}_{FR},\mathrm{V}_{SB}  \in \mathbb{R} ^{ N \times C }$). To promote effective feature collaboration, we create a cross-attention fusion operation by exchanging the query matrices $\mathrm{Q}_{FR}$ and $\mathrm{Q}_{SB}$ between the two branches as follows:

\begin{equation}\label{eq.1}
 \mathrm{CA}_{FR} = \mathrm{Softmax}\Big( \frac{\mathrm{Q}_{FR}\mathrm{K}_{SB}^{T}}{\sqrt{C}}\Big) \,\mathrm{V}_{SB},
\end{equation}

\begin{equation}\label{eq.2}
 \mathrm{CA}_{SB}= \mathrm{Softmax}\Big(  \frac{\mathrm{Q}_{SB}\mathrm{K}_{FR}^{T}}{\sqrt{C}}\Big) \,\mathrm{V}_{FR},
\end{equation}

\noindent where $\mathrm{CA}_{FR}$ and $\mathrm{CA}_{SB}$ denote the cross-attention operations and $C$ is the dimension of key matrices ($ \mathrm{K}_{FR},\mathrm{K}_{SB} \in \mathbb{R} ^{ N \times C } $). The single cross-attention operation is performed for each head in parallel to compute the multi-head cross-attention mechanism, denoted by $\mathrm{MCA}_{FR}$ and $ \mathrm{MCA}_{SB}$. Following the concatenation of all head unit outputs along the channel dimension, the resulting tensor is reshaped to match the dimensions of each feature map (${F}_{FR}, {F_{SB}} \in \mathbb{R} ^{H\times W \times C}$).

\subsubsection{Multi-scale Feed-Forward Network with Locality (MFFNL)}
The standard transformer encoder includes a feed-forward network with two fully-connected layers for up- and down-projection operations, as well as GELU \cite{hendrycks2016gaussian} activation. However, recent studies \cite{yuan2021incorporating,wang2022uformer} have shown that the vanilla feed-forward network cannot leverage local context in neighboring pixels, which is essential for an effective FR. To address this shortcoming, we propose a novel multi-scale feed-forward network named MFFNL which is able to learn facial features at different scales. As illustrated in Fig. \ref{fig:mbn}, in our proposed MFFNL block, a multi-scale depth-wise convolution (MDConv) layer is integrated into the vanilla feed-forward network. The MFFNL block consists of two pointwise convolutions for expansion and projection operations and the proposed MDConv layer is positioned in between. The MDConv layer is composed of three parallel streams, each utilizing a distinct depth-wise convolution. The first stream utilizes a $3\times3$ depthwise convolution, while the other two streams employ $1\times7$ and $7\times1$ depthwise convolutions, respectively. Motivated by \cite{szegedy2016rethinking}, decomposition is adopted to decompose a $7\times7$ convolution into two $1\times7$ and $7\times1$ convolutions to reduce computational complexity while maintaining the effective receptive field size. These streams are concatenated together to construct the fused feature representation. In addition, a shortcut connection is utilized after the MDConv layer to enhance the gradient propagation capability in the MFFNL block. The computation of the MFFNL block for input ${X}$ is represented as:
\begin{equation}
    \mathrm{MFFNL}({X}_{in}) = \mathrm{PConv}\Big(\mathrm{MDConv}\big(\mathrm{PConv}(X_{in})\big)\Big), 
\end{equation}
\begin{multline}
    \mathrm{MDConv}({X}_{in}) =  \mathrm{Concat}\Big(\mathrm{DConv}_{3\times3}({X}_{in}),\\ \mathrm{DConv}_{1\times7}({X}_{in}), 
    \mathrm{DConv}_{7\times1}({X}_{in})\Big) +{X}_{in}. 
\end{multline}

\noindent where $\mathrm{PConv}$ and $\mathrm{DConv}$ denote the pointwise and depthwise convolution layers. After each layer, we use the GELU activation and batch normalization. To sum up, the MDConv layer facilitates multi-scale feature extraction in the MFFNL block, making our CATF module able to capture both short-term and long-term dependencies in FR and SB tasks.  

\subsubsection{Channel-wise Attentional Fusion (CAF)}
 
\begin{figure}[t]
    \centering 
    \includegraphics[scale=.16]{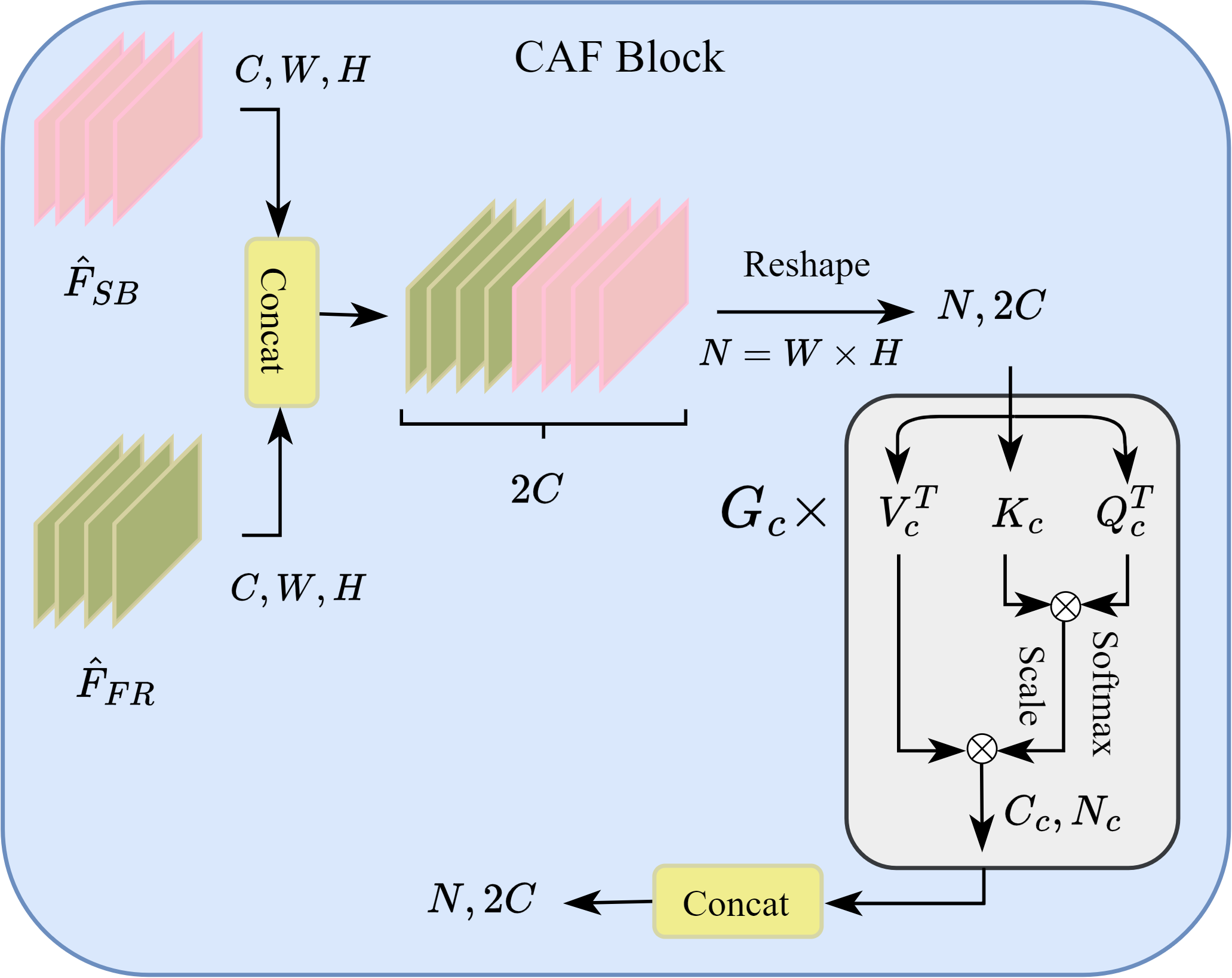}
    \caption{Proposed channel-wise attentional fusion (CAF) block.}
    \label{fig:CAF}%
\end{figure}

Once we encode the long- and short-range interactions of the FR and SB features using our cross-attention operations and the MFFNL block, we propose to conduct feature fusion across global channel tokens (see Fig. \ref{fig:CAF}). We first concatenate the outputs of MFFNL blocks, denoted by $\mathrm{{F}_{cat}} = \mathrm{Concat}({\hat{F}_{FR}},{\hat{F}_{SB}})$, along the channel dimension. Then, we construct channel-wise tokens by transposing the input tokens where the channel dimension determines the token scope and the spatial dimension determines the token feature dimension. In this context, channel-wise tokens can extract global interactions between both the FR and SB tasks. To incorporate attention scores between the channels of these tasks for feature fusion, we apply self-attention to the channel tokens. To achieve this while maintaining computational efficiency, we arrange the channel tokens into $G_c$ groups with $C_c$ channels each, where the channel dimension of $\mathrm{{F}_{cat}}$ is $2C = G_c \times C_c$. The formulation of channel-wise attention that interacts across a group of channels is as follows:
\begin{equation}\label{eq.1_}
 \mathrm{CAF}(\mathrm{Q}_{c},\mathrm{K}_{c},\mathrm{V}_{c})= \mathrm{Softmax}\Big( \frac{\mathrm{Q}^\mathrm{T}_{c}\mathrm{K}_{c}}{\sqrt{C_c}}\Big) \,\mathrm{V}^\mathrm{T}_{c},
\end{equation}

\noindent where $\mathrm{Q}_{c}, \mathrm{K}_{c}, \mathrm{V}_{c} \in \mathbb{R} ^{N_{c}\times C_c}$ are grouped channel-wise queries, keys, and values, respectively. Note that $N_{c}$ stands for the spatial dimension of each channel group. Following the projections in the DMCA module, three projection layers are employed to compute the queries, keys, and values matrices along the channel dimension. Ultimately, we calculate the $\mathrm{CAF}$ for all $G_c$ channel groups and concatenate all of them together to feed the classifier.
\section{Experiments}\label{experr}
\subsection{Implementation Details}\label{Implementation_Details}
\subsubsection{Training Datasets} \label{training_data}
To conduct a fair comparison with other methods, we separately train our model on two datasets: the CelebA \cite{celeba} and MS1MV2 \cite{webface260m} datasets. CelebA is a large-scale face dataset containing 202,559 face images from more than 10k identities with different poses, backgrounds, and lighting conditions. Each face image is annotated with 40 facial attributes such as gender, the shape of the nose, and the color of the hair. In this work, we rely on identity facial attributes that stay the same in different images of the same person. For instance, gender, the shape of eyes, or being bald remain the same in different situations such as various illuminations or poses while some attributes such as the color of hair and wearing glasses may vary in different images of the same person. In this regard, we utilize five SB attributes which are gender, big nose, chubby, narrow eye, and Bald. In line with SoTA studies, we adhere to the dataset's protocol \cite{celeba} for both the training and testing sets. The second training set, MS1MV2, is a large-scale dataset of more than 5M face images that make it possible to compare our proposed method with the SoTA methods on benchmark FR datasets. Gender information is the only attribute provided for this large-scale dataset \cite{huang2021age}. Therefore, in this case, our auxiliary modality only includes the information about one attribute. 
\begin{figure*}[ht]
    \centering 
    \includegraphics[scale=.128]{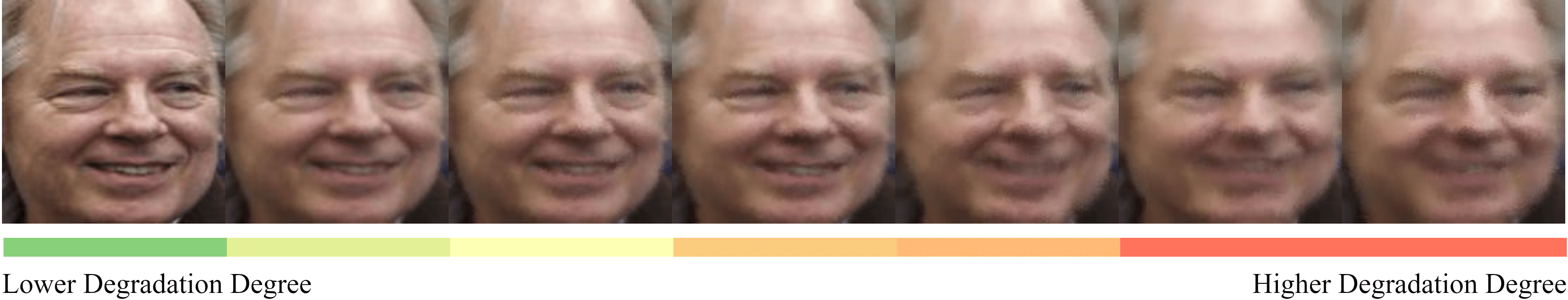}
    \caption{Images corrupted by simulated atmospheric turbulence with strengths ranging from 0.25 to 2 (the first image is the original one).}
    \label{fig:aug_turb}%
\end{figure*}

\subsubsection{Test Datasets}
To evaluate our proposed model, we utilize the test set of the CelebA dataset. Moreover, when employing the MS1MV2 dataset as the training set, we utilize several widely-used FR benchmarks with high-quality, mixed-quality, and low-quality settings. In high-quality settings, the LFW \cite{huang2008labeled}, CFP-FP \cite{CFPFP}, CPLFW \cite{CPLFWTech}, CALFW \cite{zheng2017cross}, and AgeDB \cite{agedb} datasets are utilized. The datasets in high-quality setting exhibit variations in lighting, pose, and age. To investigate the performance of the proposed method on more challenging images, we also test our model on mixed-quality setting with the IJB-B \cite{whitelam2017iarpa}, and IJB-C \cite{maze2018iarpa} datasets which cover a wide range of face variations and challenges for FR in unconstrained settings. The IJB-B and IJB-C datasets consist of 21.8K and 31.3K images, respectively. The IJB-C dataset, comprising 3,531 identities, is an extension of the IJB-B dataset, covering 1,845 different identities. For both of these datasets, we follow the standard 1:1 verification protocol which is a template-based method. Considering that each template contains multiple frames, we compute the average feature vector for each template. Moreover, to gauge the efficacy of the proposed method in more challenging scenarios, we evaluate our method on low-quality realistic and synthetic FR test sets. For realistic tests, we employ TinyFace \cite{cheng2019low}, a low-resolution in-the-wild dataset, and SCFace \cite{grgic2011scface}, a cross-resolution FR dataset captured in uncontrolled indoor environments at three different distances. As for the synthetic tests, we utilize the CelebA dataset to synthesize low-quality data corrupted by the controllable face synthesis GAN and atmospheric turbulence simulator.



\subsubsection{Augmentations}
As discussed in {Section} \ref{Proposed method}, in the training process, we utilize a novel distillation approach to transfer the knowledge learned from the high-quality images to the low-quality ones to boost the model's performance in challenging scenarios. To create realistic low-quality versions of the training data, we adopt two approaches. In our first approach, we employ a simulator proposed in \cite{mao2021accelerating} to generate images corrupted with atmospheric turbulence effect. In this simulator, we can control the strength of the atmospheric turbulence by an aperture diameter, $D$, divided by the fried parameter, $r_{0}$. We refer the readers to \cite{mao2021accelerating} for detailed information. Fig. \ref{fig:aug_turb} shows sample images degraded by different strengths of atmospheric turbulence. It is evident that when turbulence levels are high, facial attributes and landmarks are impacted by considerable deformation and blurring. During the training phase, we randomly corrupt the training data with different ratios of $D/r_{0}$ which determines the strength of the atmospheric turbulence effect (between 0.25 to 2). 

In addition to the atmospheric turbulence effect, we employ the controllable face synthesis generator introduced in \cite{liu2022controllable} to create low-quality images with unconstrained imaging environment factors including noise, low resolution, and motion blur. The generator is pre-trained on the WiderFace dataset  \cite{yang2016wider}, which encompasses a wide array of unconstrained variations, as the target data.

\subsubsection{Training details}\label{Training details}
The scale of the training dataset plays a crucial role in the performance of the FR, as a larger dataset can provide a wider range of real-world characteristics, leading to better generalization to unseen data. Hence, in the case of training on the CelebA dataset, our backbone is weighted with a pre-trained ResNet-50 \cite{he2016deep} on the VGGFace2 dataset \cite{cao2018vggface2} that contains more than 3.3M images of about 9k identities. As mentioned before, to gain a better insight into the advantage of utilizing attributes for FR, we have also used MS1MV2 \cite{webface260m}, which includes more than 5M face images as a training set. In this case, we employ ResNet-101 as the backbone of our proposed model. 

The training process of the proposed method includes two main steps. First, we jointly train our multi-branch network with both classification and distillation losses for each branch. The weight parameters of the total loss function are determined based on the prioritization of FR as our primary objective. We set $\lambda_{FR}$ = 3, $\lambda_{Male} = \lambda_{Bald} = 1$, and all other weight parameters to $0.5$. In the second step of the training phase, to enrich the FR feature representation, we fuse the FR and SB feature representations together through our proposed CATF module. As such, we train this fused branch for the goal of FR with the loss function given in Equation \ref{margin_based_softmax}.

The model undergoes training using stochastic gradient descent with an initial learning rate of 0.1. In the case of training on the CelebA dataset, the model is trained for $25$ epochs, and the scheduling step is set at $3$, $7$, and $15$ epochs. For training on the MS1MV2 dataset, the scheduling step is set at $7$, $13$, and $18$ epochs for a total of 30 epochs.

\subsection{Comparison with SoTA methods}
\subsubsection{Soft Biometric Prediction}
Table \ref{Tab:SB}, presents a comprehensive evaluation of the proposed SB predictor on the two challenging annotated datasets, CelebA \cite {celeba} and LFWA \cite{huang2008labeled} datasets. These datasets are widely used in the field of facial attribute analysis and serve as benchmarks for evaluating the performance of SB prediction methods. Similar to the SoTA methods, we have followed the same protocol, and the results of the other methods are directly reported from the original papers. In the case of the CelebA dataset, our proposed approach consistently outperformed the existing SB prediction methods, achieving the highest accuracy across all attributes. For the LFW dataset, our method surpassed all other methods for all those attributes except for the narrow eye attribute, where it secured the position of a runner-up among all other methods.

\subsubsection{Face Recognition}\label{Face Recognition}
In line with the established methodology of the SoTA FR studies that train the model as a classifier and test it as a verifier \cite{kim2022adaface, wang2018cosface, deng2019arcface}, we also evaluate our FR model as a verifier. To this end, when employing the CelebA dataset as a training set, a subset of $10,000$ pairs is randomly sampled from the CelebA dataset, ensuring that the identities of these pairs are excluded from the training set. Regarding {Section} \ref{Training details}, in the first step of the training process, we train our multi-branch network without employing any fusion modules. Thus, we can consider this branch as a baseline to better clarify the effective role of integrating SB into FR. The experimental findings presented in Table \ref{Tab:basee} provide compelling evidence that our proposed model effectively enhances FR performance through the utilization of SB attributes.
We have also performed a comprehensive set of experiments to evaluate the effectiveness of our proposed CATF module and compare it with alternative integration strategies. The experimental findings, as presented in Table \ref{Tab:basee}, demonstrate that substituting this module with simple operations like addition or concatenation leads to even inferior performance compared to the baseline approach, particularly for specific false acceptance rates (FARs). Moreover, results prove that among recent feature integration studies \cite{hu2018squeeze,zhang2022cmx,tsai2019multimodal}, our proposed module establishes the most effective integration approach for enhancing the performance of FR feature representation. Furthermore, we extend our explorations to investigate the impact of the number of attributes utilized for FR, and the last rows of Table \ref{Tab:basee} reveal that incorporating more identity facial attributes contributes to improved accuracy.

To obtain a better understanding of the benefits associated with employing SB attributes for FR and also have a fair comparison with recent SoTA methods, we have additionally trained our model on the MS1MV2 dataset \cite{deng2019arcface}. In this case, we adopt ResNet-101 as the backbone and conduct evaluations on nine widely recognized FR benchmarks. As shown in Table \ref{Tab:FRR}, results demonstrate that the observed enhancements for the mixed-quality and low-quality datasets are more notable in comparison with the improvements in the case of high-quality datasets. This can be attributed to the accuracy saturation in high-quality datasets such as LFW and CFP-FP benchmarks. As high-quality images inherently contain a wealth of important facial information, the distillation of knowledge from such images to their corresponding low-quality counterparts becomes less noticeable. Similarly, the same scenario holds true for the impact of utilizing SB attributes to help the FR branch. As a result, the marginal gains achieved in the high-quality scenario do not reflect the full potential of the proposed method. Instead, the efficacy of the proposed method becomes particularly evident when dealing with mixed-quality and low-quality images, as these cases greatly benefit from the supplementary knowledge transferred from the higher-quality images. It is worth noting that due to the availability of only one attribute for this training set, we focused on a single SB attribute (gender) to evaluate the effectiveness of the proposed method while considering multiple attributes could lead to even greater improvements in especially low-quality cases. 

To further demonstrate the effectiveness of our proposed approach, we expanded our experiments to include a realistic cross-resolution dataset. Table \ref{Tab:SCFace} presents a comparison of our proposed method with the recent SoTA methods on the SCFace dataset \cite{grgic2011scface}. Some approaches, such as RPCL \cite{li2022deep} and RI \cite{chai2023recognizability}, optimized their methods through fine-tuning on the SCFace training set. For models such as FAN \cite{yin2020fan} and TRM \cite{wang2022two}, which reported performance with and without fine-tuning on this dataset, it is evident that fine-tuning significantly enhances the performance of the FR model. As indicated in Table \ref{Tab:SCFace}, our model surpasses all non-fine-tuned methods and even rivals the performance of models that are fine-tuned on the SCFace dataset. This result underscores the generalization capability of our proposed model to handle unseen cross-resolution face images.

\begin{table}[t]

\setlength\tabcolsep{5pt}
\caption{Performance comparison in terms of accuracy (\%) between the proposed SB predictor and the SoTA methods.} 

\centering
\begin{tabular}{l | c | c c c c c} 
\hline \hline
\multirow {2}{*}{\rotatebox[origin=c]{45}{Data}} & \multirow {2}{*}{Methods} &{Bald} & Big & {Chubby} & {Male} & Narrow \\
{}&{}&{} & Nose &{}&{}&Eye\\
\hline
\multirow {7}{*}{\rotatebox[origin=c]{90}{CelebA}} & Z. Liu et al.\cite{liu2015} & 98.00 & 78.00 & 91.00 & 98.00 & 81.00\\

{} & Moon \cite{moon} & 98.77 & 84.00 & 95.44 & 98.10 & 86.52\\
{} & HyperFace \cite{ranjan2017hyperface} & - & - & - & 97.00 & -\\
{} & All-In-One \cite{ranjan2017all} & - & - & - & 99.00 & -\\
{} & MCFA \cite{zhuang2018multi} & 99.00 & 84.00 & 96.00 & 98.00 & 87.00\\
{} & DMM \cite{mao2020deep} & 99.03 & 84.78 & 95.86 & 98.29 & 87.73\\
{} & Ours & \textbf{99.11} & \textbf{85.41} & \textbf{96.13} & \textbf{99.19} & 87.69\\
\hline
\multirow {6}{*}{\rotatebox[origin=c]{90}{LFWA}} & Z. Liu et al.\cite{liu2015} & 88.00 & 81.00 & 73.00 & 94.00 & 81.00\\

{} & HyperFace \cite{ranjan2017hyperface} & - & - & - & 94.00 & -\\
{} & All-In-One \cite{ranjan2017all} & - & - & - & 93.12 & -\\
{} & MCFA \cite{zhuang2018multi} & 91.00 & 81.00 & 74.00 & 93.00 & 78.00\\
{} & DMM \cite{mao2020deep} & 91.96 & 83.67 & 77.66 & 94.14 & 83.67\\
{} & Ours & \textbf{92.19} & \textbf{84.49} & \textbf{77.71} & \textbf{95.36} & \textbf{83.81}\\
\hline \hline
\end{tabular}
 \label{Tab:SB}
\end{table}

\begin{table}[t]
\setlength\tabcolsep{5.5pt} 
\caption{Performance comparison between the proposed method (CATFace), the baseline, and other SoTA feature integration methods. Results are based on TAR$@$FAR, in which TAR and FAR stand for True Acceptance Rate, and False Acceptance Rate, respectively. Also, M, B, and NE stand for male, bald, and narrow eyes, respectively.} 
\centering
\begin{tabular}{l |c c c c c c} 
\hline \hline\rule{0pt}{2ex}  
\multirow {2}{*}{Methods}   & \multicolumn{5}{c}{CelebA} \\

\cline {2 - 6 }\rule{0pt}{3ex}    
&$10^{-5}$ & $10^{-4}$ & $10^{-3}$ & $10^{-2}$ & $10^{-1}$ \\ [1ex]
\hline
Baseline (without SB) & 89.23 & 90.51 & 92.08 & 93.26 & 94.48\\
Concatenation & 88.95 & 90.06 & 92.06 & 93.31& 94.50\\
Addition & 88.83 & 89.96 & 91.74 & 93.29 & 94.39\\
SENET \cite{hu2018squeeze}& 89.98 & 90.96 & 92.84 & 94.39 & 95.65\\
Cross-Attention \cite{tsai2019multimodal} & 90.68 & 92.12 & 93.30 & 94.50 & 95.68\\
FFM \cite{zhang2022cmx} & 90.73 & 92.59 & 93.49 & 94.52 & 95.79\\
CATFace & \textbf{91.10} & \textbf{92.91} & \textbf{93.78} & \textbf{94.83} & \textbf{96.18}\\
\cdashline{1-7}\rule{0pt}{2ex} 
CATFace (B) & 89.54 & 90.89 & 92.67 & 93.90 & 94.81\\
CATFace (B \& M) & 89.97 & 91.03 & 92.83 & 94.13 & 94.97\\
CATFace (B \& M \& NE) & 90.23 & 91.37 & 93.08 & 94.29 & 95.13\\
\hline \hline
\end{tabular}
\label{Tab:basee}
\end{table}

\begin{table*}[t]
\label{9test}
\setlength\tabcolsep{9.5pt}
\caption{Performance comparison of our proposed method (CATFace) with recent SoTA FR methods. TAR is reported at FAR = $0.01\%$ (All these methods are trained on the MS1MV2 dataset).} 
\vspace{2mm}
\centering
\begin{tabular}{l | c | c c c c c | c c | c c} 

\hline \hline
\multirow{3}{*}{Methods}&\multirow{3}{*}{Venue} &\multicolumn{5}{c|}{High Quality} & \multicolumn{2}{c|}{Mixed Quality}& \multicolumn{2}{c}{Low Quality}  \\\cline {10- 11}&
{} &\multicolumn{5}{c|}{(Verification Accuracy)} & \multicolumn{2}{c|}{(TAR)}& \multicolumn{2}{c}{TinyFace}  \\ \cline { 3 - 11}\rule{0pt}{2ex} & {}& LFW & CFP-FP & CPLFW & AgeDB & CALFW & IJB-B& IJB-C & Rank-1 & Rank-5\\ 
\hline
CosFace\cite{wang2018cosface} &CVPR18 & 99.81 & 98.12 & 92.28 & 98.11 & 95.76 &94.80 & 96.37& - & - \\
ArcFace\cite{deng2019arcface} &CVPR19 & 99.83 & 98.27 & 92.08 & 98.28 & 95.45 &94.25 & 96.03 & - & -  \\
MV-Softmax\cite{wang2020mis} &AAAI20 & 99.80 & 98.28 & 92.83 & 97.95 & 96.10 &93.60 & 95.20 & - & - \\
URL\cite{shi2020towards} &CVPR20 & 99.78 & 98.64 & - & - & - & - & 96.60 & 63.89 & 68.67 \\
SCF-ArcFace\cite{9577756} &CVPR21 & 99.82 & 98.40 & 93.16 & 98.30 & 96.12 & 94.74 & 96.09 & - & - \\
MagFace\cite{meng2021magface} &CVPR21 & 99.83 & 98.46 & 92.87 & 98.17 & 96.15 & 94.51 & 95.97 & - & - \\
MIND\cite{low2021mind}&LSP21 & - & - & - & - & - & - & - & 66.82 & -\\
ElasticFace\cite{boutros2022elasticface}& CVPRW22 & 99.80 & \textbf{98.73} & 93.23 & 98.28 & 96.18 & 95.43 & 96.65 & - & -\\
LS\cite{wang2023low}&FG23 &99.50 & - & - & - & - & - & - & 66.30 & -\\
\cdashline{1-11}\rule{0pt}{2ex} 
AdaFace\cite{kim2022adaface} &CVPR22 & 99.82 & 98.49 & 93.53 & 98.05 & 96.08 & 95.67 & 96.89 & 68.21 & 71.54 \\
CATFace$^{1}$ & {-}& 99.83 & 98.57 & \textbf{93.71} & \textbf{98.14} & \textbf{96.17} & \textbf{95.82} &\textbf{97.07} & \textbf{68.52} &\textbf{71.92}  \\
CATFace$^{2}$ & {-}& \textbf{99.84} & 98.68 & \textbf{93.84} & \textbf{98.33} & \textbf{96.32} & \textbf{96.13} &\textbf{97.43} & \textbf{68.95} &\textbf{72.31}  \\
\hline \hline
\end{tabular}
\begin{tablenotes}
      \small
      \item $^{1}$This is our proposed FR method trained with the self-distillation approach without employing SB attributes.
     \item $^{2}$This is our proposed FR method trained with both the self-distillation approach and the proposed CATF module to employ SB attributes.
    \end{tablenotes}
  \label{Tab:FRR}
\end{table*}

\begin{table*}[ht] \label{Ablation_self_Distillation}

\setlength\tabcolsep{12.3pt}
\caption{Ablation of our self-attention distillation approach on the SB branch.} 

\centering
\begin{tabular}{l | c c c c| c c c c c} 
\hline \hline
\multirow {3}{*}{Test Data} & \multicolumn{4}{c|}{Approach}& \multirow {3}{*}{Bald} & \multirow {3}{*}{Big Nose} & \multirow {3}{*}{Chubby} & \multirow {3}{*}{Male} & \multirow {3}{*}{Narrow Eye} \\

\cline {2 - 5 }& \multirow {2}{*}{Aug} & \multicolumn{3}{|c|}{Distill} & {} & {}& {}& {}& {} \\
\cline {3 - 5 }&{}& \multicolumn{1}{|c}{Feat} & CBAM  & SA & {} & {}& {}& {}& {}\\
\hline
\multirow {5}{*}{CelebA} & {} & {} & {} & {}& 99.10 & 84.84 & 96.09 & 99.16 & 87.56\\
{} & $\checkmark$ & {} &{}& {} & 99.10 & 84.95 & 96.00 & 99.10 & 87.30\\
{} & $\checkmark$ & $\checkmark$ &{}& {} & 99.11 & 84.91 & 96.05 & 99.14 & 87.57\\
{} & $\checkmark$ & {}&$\checkmark$& {} &  99.12 & 85.16 & 96.10 & 99.16 & 87.63\\
{} & $\checkmark$ & {} & {} & $\checkmark$ &  99.11 & \textbf{85.41} & \textbf{96.13} & \textbf{99.19} & \textbf{87.69}\\
\hline
\multirow {5}{*}{Distorted CelebA} & {} & {} &{}& {} & 96.53 & 80.73 & 93.84 & 97.68 & 84.79\\
{} & $\checkmark$ & {} &{}& {} &  97.70 & 82.01 & 94.24 & 97.93 & 85.32\\
{} & $\checkmark$ & $\checkmark$ &{}& {} &  98.09 & 82.49 & 94.30 & 97.96 & 85.68\\
{} & $\checkmark$ & {} &$\checkmark$& {} &  98.20 & 83.05 & 94.54 & 98.41 & 85.93\\
{} & $\checkmark$ & {} & {} & $\checkmark$ &  \textbf{98.31} & \textbf{83.11} & \textbf{94.69} & \textbf{98.74} & \textbf{86.10}\\
\hline
\multirow {5}{*} {LFWA} &  {} & {} & {} & {} &  90.83 & 82.73 & 76.67 & 93.10 & 82.96\\
{} & $\checkmark$ & {} &{}& {} &  90.96 & 82.97 & 76.83 & 93.13 & 82.95\\
{} & $\checkmark$ & $\checkmark$ &{}& {} &  91.23 & 82.98 & 76.91 & 93.45 & 83.02\\
{} & $\checkmark$ & {} &$\checkmark$& {} &  91.87 & 83.07 & 77.39 & 94.09 & 83.85\\
{} & $\checkmark$ &{} &  {} & $\checkmark$ &  \textbf{92.19} & \textbf{84.49} & \textbf{77.71} & \textbf{95.36} & \textbf{83.81}\\
\hline
\multirow {5}{*} {Distorted LFWA} &   {} & {} & {} & {}&  88.00 & 78.12 & 72.22 & 90.62 & 77.38\\
{} & $\checkmark$ & {} &{}& {} & 88.91 & 79.35 & 73.10 & 90.99 & 78.45\\
{} & $\checkmark$ & $\checkmark$ &{}& {} &  89.15 & 79.43 & 73.93 & 91.75 & 78.90\\
{} & $\checkmark$ & {} &$\checkmark$& {} &  89.88 & 80.91 & 74.50 & 92.90 & 79.25\\
{} & $\checkmark$ & {} &{}& $\checkmark$ &  \textbf{90.26} & \textbf{81.51} & \textbf{74.66} & \textbf{93.02} & \textbf{80.62}\\
\hline \hline
\end{tabular}
 \label{Tab:SB_disill}
\end{table*}

\subsection{Ablation and Analysis}
\subsubsection{Effect of Self-Attention Distillation}
To better clarify the effect of self-attention distillation on our proposed model, we investigate its impact on the FR and SB branches separately. Table \ref{Tab:SB_disill} and Table \ref{Tab:FR_disill} demonstrate the effect of self-attention distillation on SB prediction and FR, respectively. In the training phase, to generate distorted versions of both CelebA and LFWA datasets, within each batch, we randomly distort 50 percent of images by atmospheric turbulence \cite{mao2021accelerating} and the rest are augmented by a GAN-based controllable face synthesis method \cite{liu2022controllable}. For each set of test data, we have considered five different models to predict SB attributes. The first model is the SB branch of a simple multi-branch network trained without any data augmentation and distillation approaches (the first row of Table \ref{Tab:SB_disill}). The second model is trained with augmented data, in addition to the original data without employing any distillation approach. The third model utilizes a general feature-based distillation method during the training phase. The fourth and last models use attention-based self-distillation approaches which are based on CBAM \cite{woo2018cbam} and self-attention, respectively. The results in Table \ref{Tab:SB_disill} indicate that while data augmentation can improve the accuracy of predicting certain SB attributes, the incorporation of KD proves to be more effective in fully utilizing the potential of the synthesized low-quality images. Furthermore, our experimental findings verify the superiority of the self-attention mechanism in producing informative feature maps in comparison with the CBAM method. Similarly, for the FR task, we have considered five different models (see Table \ref{Tab:FR_disill}). The first model corresponds to the FR branch of a simple multi-branch network trained without data augmentation or distillation approaches. The other four models are also similar to what was mentioned for the SB branch. Results obtained from these experiments for the FR task further reinforce the effectiveness of our proposed KD approach over alternative methods in KD.

\begin{table}[t]

\setlength\tabcolsep{6.5pt} 
\caption{Performance comparison of our proposed method (CATFace) with recent SoTA FR methods on the SCFace dataset.} 
\centering
\begin{tabular}{c |c| c c c |c} 
\hline \hline
\multirow {2}{*}{Methods}& \multirow {2}{*}{Fine-Tuned}& \multicolumn {3}{c|}{Distance} & \multirow {2}{*}{Avg.} \\

\cline {3 - 5}\rule{0pt}{2ex}

{} & {} &$4m$ & $2.6m$ & \multicolumn {1}{c|}{$1m$} & {}  \\
\hline\rule{0pt}{2ex} 

FAN\cite{yin2020fan} & $\checkmark$   &77.50 & 95.00& 98.30& 90.30 \\
ArcFace\cite{deng2019arcface,soni2022synthetic} & $\checkmark$ & 80.50 & 98.00 & 99.50 & 92.70 \\
RPCL\cite{li2022deep} & $\checkmark$  & 90.40& 98.00& 98.00 &95.47 \\

TRM\cite{wang2022two} & $\checkmark$   &91.25& 99.50& 99.50 & 96.75 \\
DDL\cite{huang2020improving} & $\checkmark$   & 93.20 & 99.20& 98.50 & 97.00 \\
RI\cite{chai2023recognizability} &$\checkmark$ & 97.07 & 99.23& 99.80 &98.70 \\
\cdashline{1-6}\rule{0pt}{2ex} 
FAN\cite{yin2020fan} & {}  & 62.00 & 90.00  & 94.80 & 82.30 \\
ArcFace\cite{deng2019arcface,soni2022synthetic} & {}  & 58.90 & 98.30 & 99.50 & 85.50 \\
DCR\cite{lu2018deep} & {} & 73.30 & 93.50 & 98.00 &  88.27 \\
TRM\cite{wang2022two} & {}   & 79.25 & 97.00 & 97.75  & 91.33 \\
CATFace & {} &  \textbf{90.64} & \textbf{98.85}& \textbf{99.61}  &  \textbf{96.37} \\

\hline \hline
\end{tabular}
  \label{Tab:SCFace}
\end{table}

\begin{figure}[t] 
    \centering 
    \includegraphics[scale=0.155]{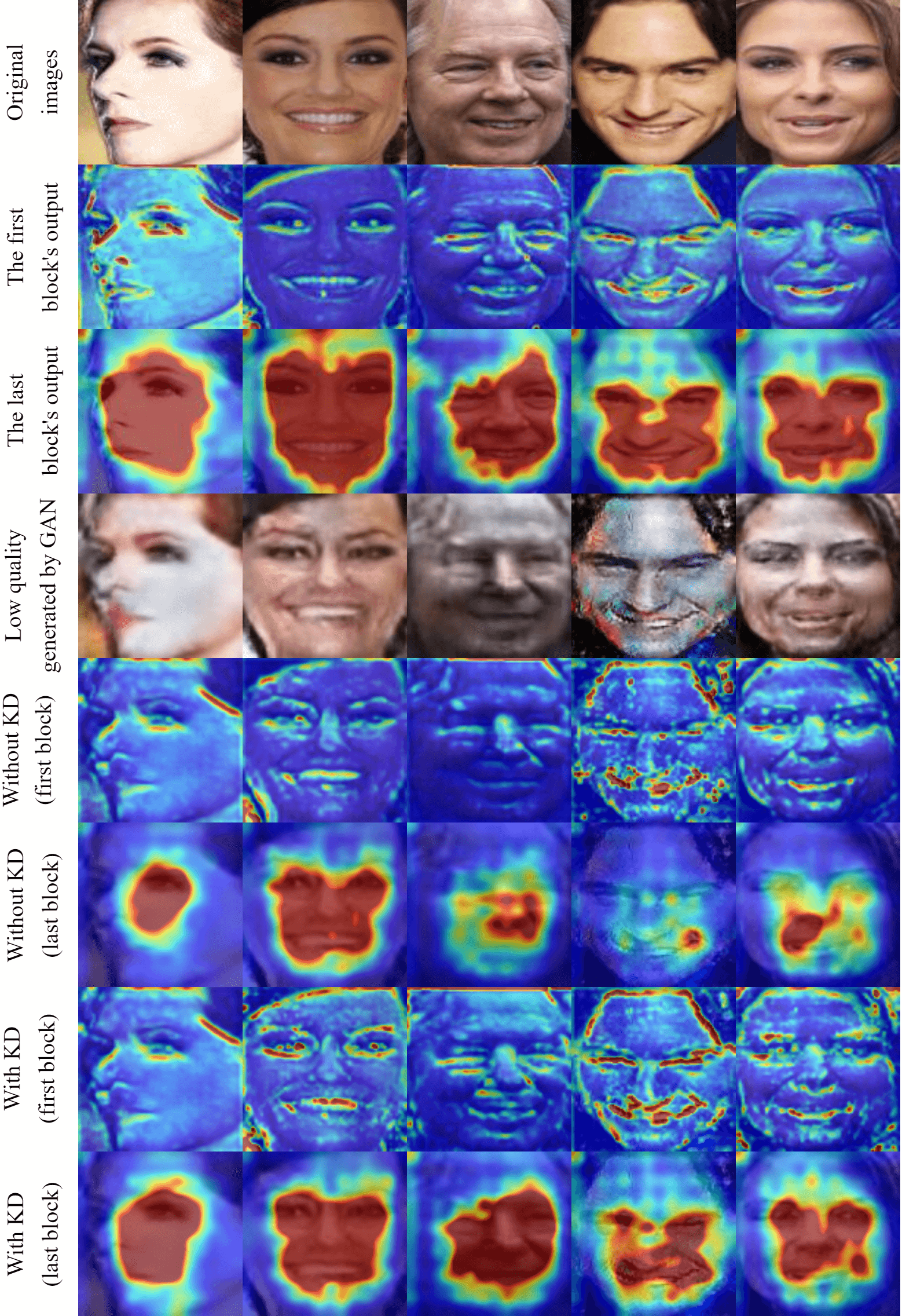}
    \caption{The visualization of the feature maps of the first and the last block of the FR branch. The second and the third rows are related to the original images. The fourth row shows the low-quality versions of the original images generated by the GAN. The fifth and the sixth rows are corresponding to the low-quality images without using the KD approach while the last two rows depict the feature maps of the low-quality images when the network is utilizing the proposed KD approach.} 

    \label{fig:attention_CFSM}%
\end{figure}

\begin{table}[!h] \label{Ablation_self_Distillation2}

\setlength\tabcolsep{7.6pt}
\caption{Ablation of our self-attention distillation approach on the FR branch (TAR is reported at FAR = 0.01$\%$).} 

\centering
\begin{tabular}{ c c c c| c c} 
\hline \hline
\multicolumn{4}{c|}{Approach}& \multirow {3}{*}{CelebA} & \multirow {3}{*}{Distorted CelebA} \\

\cline {1 - 4 }\rule{0pt}{2ex}  \multirow {2}{*}{Aug} & \multicolumn{3}{|c|}{Distill} & {} & {} \\

\cline {2- 4 }\rule{0pt}{2ex} & \multicolumn{1}{|c}{Feat} & CBAM & SA & {} & {}\\

\hline
 {} & {} & {}&{}& 92.64 & 89.16\\
 $\checkmark$ & {} &{}& {} & 92.83 & 90.36\\
 $\checkmark$ & $\checkmark$ &{}& {} & 93.05 & 90.73\\
 $\checkmark$ & {}&$\checkmark$& {} &  93.18 & 91.27\\
 $\checkmark$ & {} & {} & $\checkmark$ &  \textbf{93.26} & \textbf{91.43}\\

\hline \hline
\end{tabular}
 \label{Tab:FR_disill}
\end{table}

\begin{figure}[t]
    \centering 
    \includegraphics[scale=0.155]{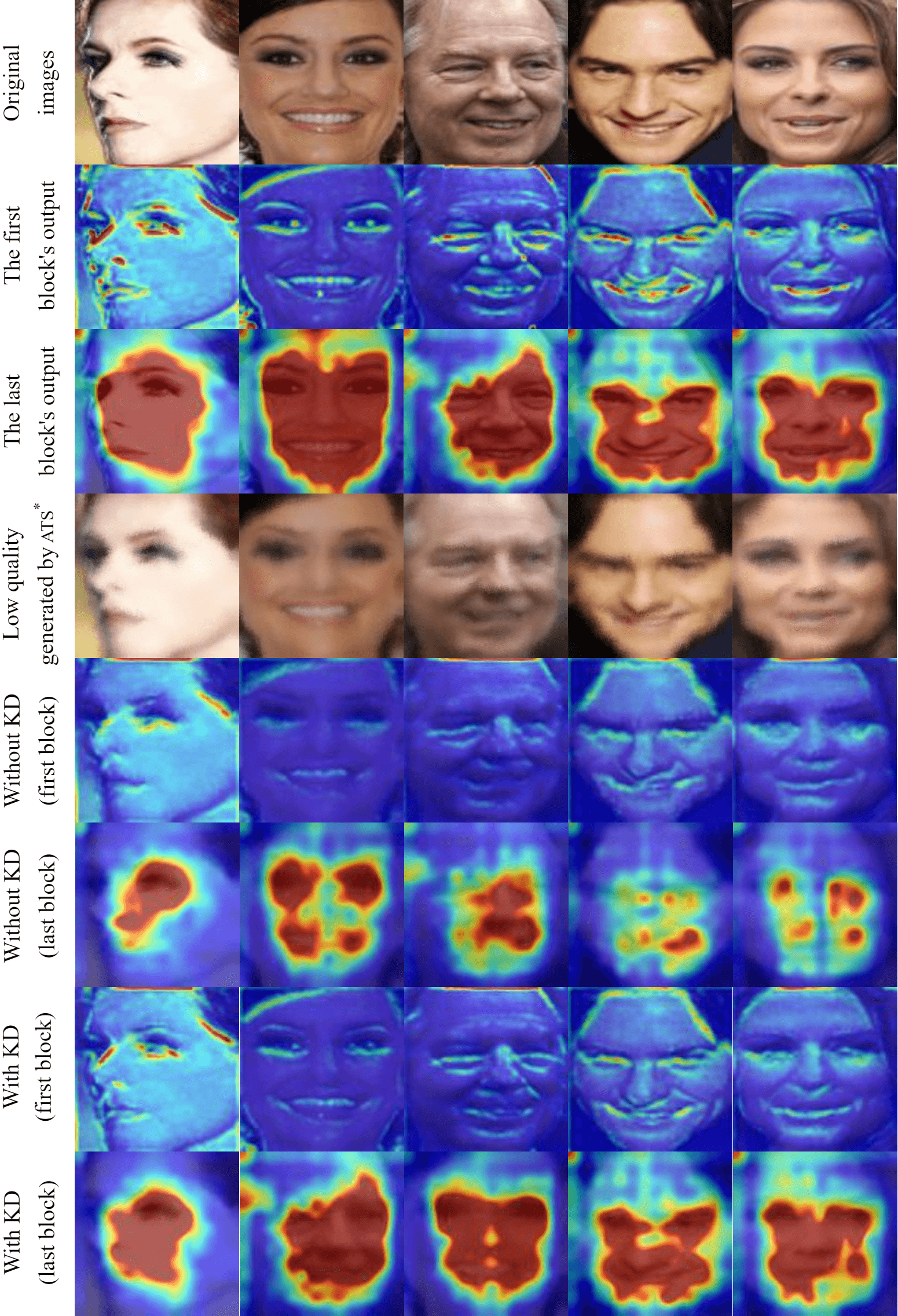}
    \caption{The visualization of the feature maps of the first and the last block of the FR branch. The second and the third rows are related to the original images. The fourth row shows the low-quality versions of the original images which are disturbed by atmospheric turbulence effects. The fifth and the sixth rows are corresponding to the low-quality images without using the KD approach while the last two rows depict the feature maps of the low-quality images when the network is utilizing the proposed KD approach. \\
    ${}^\star$ATS stands for atmospheric turbulence simulator.}
    \label{fig:attention_Atmospheric}%
\end{figure}

\subsubsection{Effect of Cosine Similarity metric on Distillation} \label{cosine_norm}
Table \ref{Tab:cosine} compares four different approaches, each varying in attention map creation and distillation metrics. The first two rows utilize the L2-distance metric to distill attention maps from high-quality samples to their corresponding low-quality counterparts. The subsequent rows adopt a similar approach but employ the cosine similarity metric for attention distillation. Results conspicuously verify that utilizing cosine similarity, whether with CBAM or self-attention methods, outperforms using L2-distance for distillation. It is demonstrated that feature norm is positively correlated with image quality \cite{meng2021magface, kim2022adaface}. Thus, to effectively utilize the information extracted from the high-quality samples, we need to distill only the directional component of the attention maps while the L2-distance tries to align both the norm and angle components of the attention maps. As shown in Equations \ref{loss_distill1} and \ref{loss_distill2}, our distillation loss minimizes exclusively the angle between the attention maps of the high-quality and low-quality samples which enables the network to focus on richer information.

\subsubsection{Effect of Soft Biometric Attributes}
As explained in {Section} \ref{Face Recognition}, to explore the contribution of SB attributes to our proposed model, we conduct several experiments such as exploring the impact of the number of attributes employed for FR or clarifying the crucial role of the proposed CATF module in the integration stage. All the experimental results verify the benefits of serving SB information as auxiliary data in the FR task (see Tabel \ref{Tab:basee}). In this section, we perform further experiments to assess the contribution of each component of our proposed CATF module, namely MFFNL and CAF blocks, to the FR task. Table \ref{Tab:ablation_catf} demonstrates that optimal performance is attained when both components are utilized in conjunction. Furthermore, the CAF block appears to have a more significant impact on the overall performance of the CATF module compared to the MFFNL block. 

\begin{figure}[t]
    \centering 
    \includegraphics[scale=0.0443]{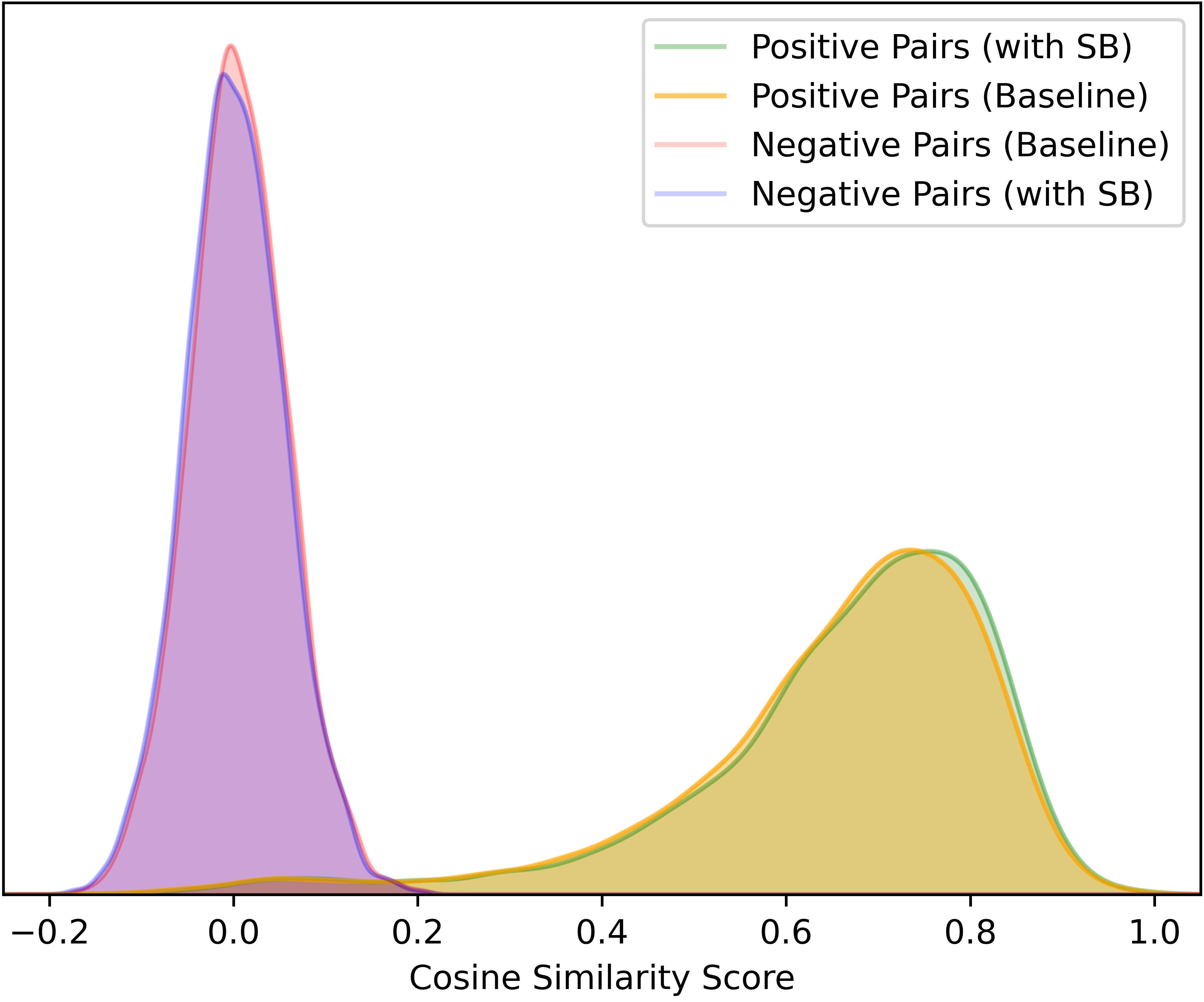}
    \caption{Comparison of cosine similarity distribution between the baseline model without employing SB information and the proposed model employing SB information. The positive and negative pairs are sampled from the CelebA dataset.}
    \label{fig:distribution}%
\end{figure}

To gain deeper insights into the role of SB attributes, we visualize the distributions of the similarity scores on 10,000 pairs of the CelebA dataset both with and without the utilization of SB attributes. As depicted in Fig. \ref{fig:distribution}, the peak values of the cosine similarity score distribution for both the positive and negative pairs are shifted. To be more specific, the peak value of the cosine similarity score distribution is shifted rightward for the positive pairs and leftward for the negative pairs. These shifts indicate that leveraging SB attributes leads to better separation between the similarity scores of the positive and negative pairs which implies a reduction in false positive and false negative errors.







\begin{table}[t] \label{Ablation_cosine}

\setlength\tabcolsep{4.9pt}
\caption{Ablation of our distillation metric on the FR branch (TAR is reported at FAR = 0.01$\%$).} 

\centering
\begin{tabular}{ c c c c| c c}
\hline \hline
\multicolumn{4}{c|}{Approach}& \multirow {3}{*}{CelebA} & \multirow {3}{*}{Distorted CelebA} \\

\cline {1 - 4 } \multicolumn {2}{c}{Distance Metric} & \multicolumn{2}{|c|}{Distill} & {} & {} \\

\cline {1-4 } \multicolumn{1}{c}{L2-Distance} & Cosine Sim & \multicolumn{1}{|c}{CBAM} & SA &{} & {}\\

\hline
$\checkmark$ & {} &\multicolumn{1}{|c}{$\checkmark$} & {} & 93.07 &90.77\\
$\checkmark$ & {} &\multicolumn{1}{|c}{}& $\checkmark$ & 93.10 & 90.89\\
{} & $\checkmark$ & \multicolumn{1}{|c}{$\checkmark$}& {} & 93.18 & 91.27\\
{} & $\checkmark$ & \multicolumn{1}{|c}{} & $\checkmark$ & \textbf{93.26} &\textbf{91.43}\\

\hline \hline
\end{tabular}
 \label{Tab:cosine}
\end{table}

\begin{table}[t]

\setlength\tabcolsep{8.4pt} 
\caption{Ablation of our proposed CATF module on the MFFNL and CAF blocks (results are based on TAR@FAR).} 
\centering
\begin{tabular}{c c| c c c c c} 
\hline \hline
\multicolumn{2}{c|}{Approach}& \multicolumn {5}{c}{CelebA}  \\

\hline

MFFNL & CAF &$10^{-5}$ & $10^{-4}$ & $10^{-3}$ & $10^{-2}$ & $10^{-1}$ \\

\hline
{}&{}& 90.68 & 92.12 & 93.30 & 94.50 & 95.68\\
$\checkmark$& {} & 90.81 & 92.76& 93.55& 94.73 & 95.90\\
{}& $\checkmark$ & 90.97 & 92.83& 93.72& 94.78 & 96.04\\
$\checkmark$&$\checkmark$ &  \textbf{91.10} & \textbf{92.91} & \textbf{93.78} & \textbf{94.83} & \textbf{96.18}\\

\hline \hline
\end{tabular}
  \label{Tab:ablation_catf}
\end{table}

\subsubsection{Visualization}
The features of the first and last convolutional block of the FR branch are visualized in Fig. \ref{fig:attention_CFSM} and Fig. \ref{fig:attention_Atmospheric}, through an attention map introduced by \cite{zagoruyko2016paying}. We compare the output feature representations between the original data and its low-quality version generated by the controllable GAN and the atmospheric turbulence simulator. To generate these maps, we normalize values within a range of 0 to 1, making them more visually interpretable. The attention maps provide valuable insights into the network's focus during FR tasks. Regarding these attention maps, in the case of high-quality images (the original data), the network focuses on critical facial features such as the eyes, nose, and lips, which play a pivotal role in ensuring accurate recognition (rows 2 and 3, Fig. \ref{fig:attention_CFSM} and Fig. \ref{fig:attention_Atmospheric}). However, when it comes to the low-quality images, the network's attention to intricate details such as facial landmarks is compromised due to the absence of information (rows 5 and 6, Fig. \ref{fig:attention_CFSM} and Fig. \ref{fig:attention_Atmospheric}).
This is where our attention-based KD approach comes into play and enhances the ability of the model to concentrate on detailed facial features (the last two rows, Fig. \ref{fig:attention_CFSM} and Fig. \ref{fig:attention_Atmospheric}). It achieves this by matching the attention maps of the low-quality images to their corresponding high-quality counterparts.

\section{Conclusion}
This paper addresses the poor performance of FR models with regard to low-quality images. Inspired by the fact that humans intrinsically analyze facial attributes to recognize identities, we utilize SB attributes as auxiliary information to improve the performance of FR. We propose a novel feature-level fusion module to effectively integrate SB information into the FR feature representations. We also incorporate a self-distillation technique during the simultaneous training of both the SB and FR branches which empowers our network to extract and distill effective information from high-quality samples, thereby strengthening its capacity to handle low-quality input images.


%

\ifCLASSOPTIONcaptionsoff
  \newpage
\fi



\bibliographystyle{IEEEtran}
%

\bibliography{arxive.bib}

%

\vskip -2\baselineskip plus -1fil

\begin{IEEEbiography}[{\includegraphics[width=1.in,height=1.25in,clip,keepaspectratio]{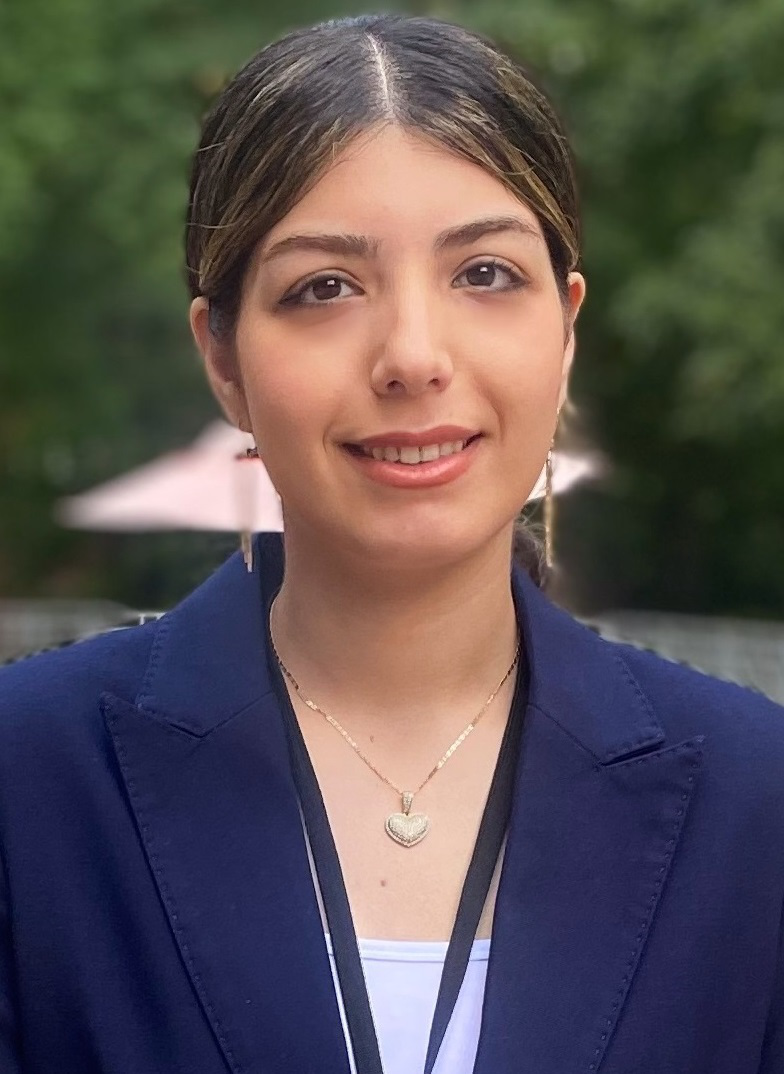}}]{Niloufar Alipour Talemi}
received the B.Sc. degree in electrical engineering from Shahid Beheshti University, Tehran, Iran, and the M.Sc. degree in electrical engineering digital electronic systems from University of Guilan, Rasht, Iran. She is currently pursuing the Ph.D. degree with West Virginia University,
USA. Her current research interests include computer vision, pattern recognition, deep learning, machine learning, and their applications in biometrics.  

\end{IEEEbiography}

\vskip -2.9\baselineskip plus -1fil

\begin{IEEEbiography}[{\includegraphics[width=1in,height=1.25in,clip,keepaspectratio]{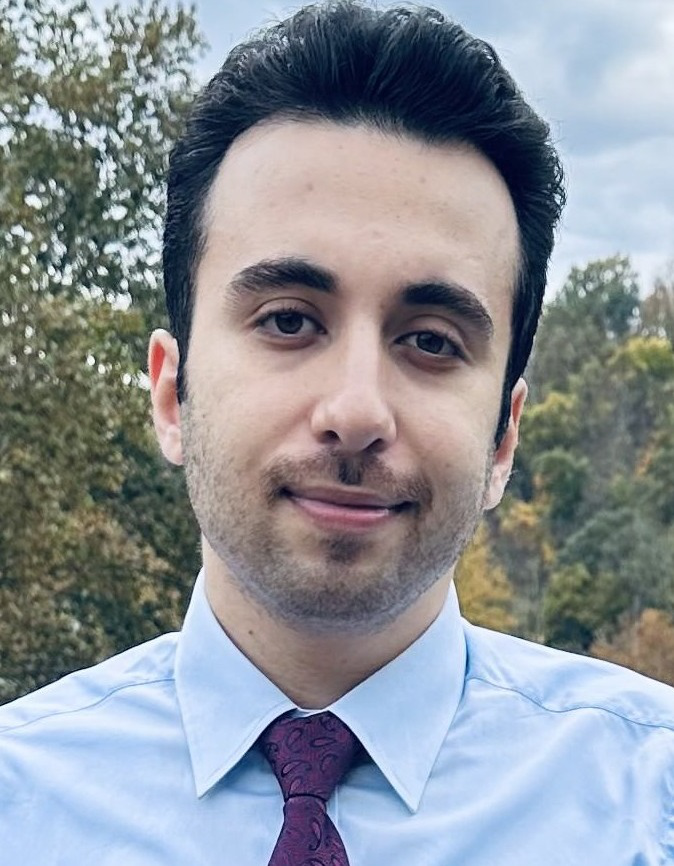}}]{Hossein Kashiani} received the M.Sc. degree in electrical engineering from Iran University of Science and Technology, Tehran, Iran. Currently, he is pursuing the Ph.D. degree at West Virginia University, USA. He has authored over 15 publications, including journals and peer-reviewed conferences. His current research interests encompass computer vision, deep learning, machine learning, and their applications in biometrics. Furthermore, he has served as a reviewer for \textsf{IEEE Transactions on Circuits and Systems for Video Technology, Pattern Recognition, Expert Systems With Applications, Knowledge-Based Systems}, and other related journals and conferences.
\end{IEEEbiography}

\vskip -2.7\baselineskip plus -1fil

\begin{IEEEbiography}[{\includegraphics[width=1in,height=1.25 in,clip,keepaspectratio]{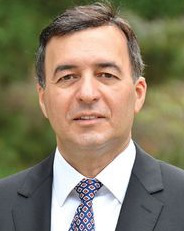}}]{Nasser M. Nasrabadi}
(Fellow, IEEE) received the B.Sc. (Eng.) and Ph.D. degrees in electrical engineering from the Imperial College of Science and Technology, University of London, London, U.K., in 1980 and 1984, respectively. In 1984, he was with IBM, U.K., as a Senior Programmer.
From 1985 to 1986, he was with Philips Research Laboratory, New York, NY, USA, as a member of the Technical Staff. From 1986 to 1991, he was an Assistant Professor with the Department of Electrical Engineering, Worcester Polytechnic Institute, Worcester, MA, USA. From 1991 to 1996, he was an Associate Professor with the Department of Electrical and Computer Engineering, State University of New York at Buffalo, Buffalo, NY, USA. From 1996 to 2015, he was a Senior Research Scientist with the U.S. Army Research Laboratory. Since 2015, he has been a Professor with the Lane Department of Computer Science and Electrical Engineering. His current research interests are in image processing, computer vision, biometrics, statistical machine learning theory, sparsity, robotics, and neural networks applications to image processing. He has served as an Associate Editor for the \textsf{IEEE Transactions on Image Processing, IEEE Transactions on Circuits and Systems for Video Technology,} and the \textsf{IEEE Transactions on Neural Networks}. He is a Fellow of \textsf{ARL} and \textsf{SPIE}.
\end{IEEEbiography}




\end{document}